\documentclass[letterpaper, 10 pt, conference]{config/ieeeconf}  %

\IEEEoverridecommandlockouts                  

\usepackage{etoolbox}
\makeatletter 
\let\NAT@parse\undefined
\makeatother
\usepackage[numbers]{natbib}
\usepackage[bookmarks=true, colorlinks=false,
    linkcolor=black,
    filecolor=mrobota,
    urlcolor=cyan]{hyperref}
\usepackage{amsmath}
\usepackage{amsfonts, amssymb}
\usepackage{algorithm}
\usepackage{soul}
\usepackage{algorithmic}
\usepackage[dvipsnames]{xcolor}
\usepackage[font=small,skip=3pt]{caption}
\usepackage{subcaption}
\usepackage{adjustbox}
\usepackage{booktabs}
\usepackage{graphicx}
\usepackage{csvsimple}
\usepackage{tikz}
\usepackage{pgf}
\usepackage{import}
\usepackage{pgfplots}
\usepackage{pgfplotstable}
\usepgfplotslibrary{fillbetween}
\usepackage{multirow}
\usepackage{siunitx}
\usepackage{svg}

\usepackage{graphicx}
\usepackage{booktabs}
\usepackage{multirow}
\usepackage{array}
\usepackage{makecell}
\usepackage{adjustbox}

\usepackage[utf8]{inputenc}

\newtheorem{problem}{Problem}

\usepackage{stackengine}
\usepackage{color}
\usepackage{todonotes}
\usepackage{amssymb}
\usepackage{mathtools}
\usepackage{orcidlink}
\newcommand{\noboxorcid}[1]{%
  {\hypersetup{pdfborder={0 0 0}}\href{https://orcid.org/#1}{\orcidlink{#1}}}%
}

\usepackage{comment}
\makeatletter
\title{\Large \textbf{BOWConnect}: Parallel Bayesian Optimization over Windows with Learned Local Cost Maps for Sample-Efficient Kinodynamic Motion Planning}

\author{{Sourav Raxit$^{1}$\noboxorcid{0000-0003-1196-2435}}, {Abdullah Al Redwan Newaz$^{1^*}$\noboxorcid{0000-0003-1140-8119}}, {Jose Fuentes$^{2}$\noboxorcid{0000-0002-6477-5820}}, and {Leonardo Bobadilla$^{2}$\noboxorcid{0000-0003-2097-2432}}
\thanks{
This work is supported in part by the U.S. EPA grant BR-02F47801-5010M, NSF grants 2118329, IIS-2024733, IIS-2331908, ONR grant N00014-23-1-2789, DoD grant 78170-RT-REP, the ARL under contract W911NF1920243, and the FDEP grant INV31.}
\thanks{$^{1}$ S. Raxit, and A. A. R. Newaz (* corresponding author) are with the Department of Computer Science, Louisiana State University New Orleans, New Orleans, LA 70148, USA (email: \{sraxit, aredwann\}@uno.edu).
$^{2}$ J. Fuentes, and L. Bobadilla are with the Knight Foundation School of Computing and Information Sciences, Florida International University, Miami, FL 33199, USA (email:%
        \{jfuen099@, bobadilla@cs.\}fiu.edu).
        }
}
\begin{document}
    \maketitle
    \begin{abstract}
 This paper presents BOWConnect, a bidirectional parallel kinodynamic motion planner that addresses three fundamental limitations of existing sampling-based methods: sample inefficiency in high-dimensional state spaces, unreliable cost heuristics under dynamic constraints, and poor performance in narrow passage environments. Unlike classical planners that rely on random control sampling and geometric distance heuristics, BOWConnect integrates Bayesian Optimization over Windows (BOW) as a learning-based steering function within a parallel tree-based exploration framework, enabling each worker to learn local cost maps and constraints to guide sampling toward dynamically feasible and collision-free controls. A bidirectional architecture simultaneously grows forward and backward trees from the start and goal regions in parallel threads, with a spatial hashing mechanism enabling fast connection queries and a boundary value problem solver generating kinodynamically consistent bridge trajectories. Extensive evaluations across ten benchmark environments demonstrate that BOWConnect achieves 100\% success while delivering the fastest or near-fastest planning time in complex scenarios, including narrow passages and non-convex spaces where state-of-the-art planners fail or degrade substantially. Real-world deployment on a ground vehicle and a quadrotor confirms real-time planning with no collisions. Videos of real-world and simulated experiments, high-resolution
versions of the figures, and the open-source code are available at
\url{https://bow-connect.github.io/}.
\end{abstract}

    \section{Introduction}
\label{sec:introduction}

Kinodynamic motion planning remains a fundamental challenge in robotics because 
the relevant search space is the system \emph{state space}, which is typically 
higher-dimensional than configuration space due to the inclusion of configuration variables and their derivatives (e.g., velocities and higher-order 
derivatives)~\cite{lavalle2001randomized,hsu2002randomized}. This dimensionality 
makes sampling-based exploration susceptible to the curse of dimensionality and 
slow convergence~\cite{meng2024online,zheng2021accelerating}.

Beyond dimensionality, kinodynamic planning must satisfy differential constraints 
throughout the trajectory~\cite{lavalle2001randomized,hsu2002randomized}, meaning 
geometrically collision-free paths may still be dynamically infeasible. For many 
systems, an exact steering function is difficult or impractical to 
derive~\cite{li2016asymptotically}, so planners typically generate motion by 
sampling controls and forward-propagating the dynamics rather than by simple 
interpolation~\cite{hsu2002randomized,csucan2009kinodynamic,kleinbort2018probabilistic}. 
Guiding this search is equally difficult: existing methods often rely on 
Euclidean distance heuristics for candidate selection because estimating 
transition costs under dynamics is non-trivial, though learning-based approaches 
have recently been proposed to address this 
limitation~\cite{lai2024neural,faroni2023motion}. These challenges compound 
further in highly constrained environments such as narrow passages, where 
generating dynamically feasible connections through constrained regions is 
particularly demanding~\cite{csucan2009kinodynamic,boeuf2015enhancing}.

Parallelism has improved performance for geometric planners such as RRT-Connect 
through GPU-parallel expansion and collision 
checking~\cite{huang2025prrtc}, but kinodynamic planners have historically been 
harder to parallelize due to propagation-dominated, serial computation patterns. 
This has motivated recent work that explicitly restructures kinodynamic tree 
growth for massively parallel devices~\cite{perrault2025kino}, yet scalable 
bidirectional kinodynamic planning with efficient connection discovery remains an 
open problem.

In this work, we present \textbf{BOWConnect}, a bidirectional and parallel 
kinodynamic motion planner designed to improve sample efficiency and accelerate 
connection discovery in continuous state spaces. BOWConnect grows two trees 
simultaneously from the start and goal, using an online learning-based trajectory 
generator to produce collision-free, kinodynamically feasible motions directly in 
continuous state and control spaces. A spatial hashing mechanism embedded in the 
\texttt{MotionTree} data structure enables efficient multi-stage feasibility 
verification, reducing redundant propagation and accelerating connection queries 
between the two trees. Together, these components improve computational efficiency 
while maintaining trajectory feasibility across diverse robotic platforms, as 
demonstrated through experiments on ground and aerial vehicles.

The main contributions of this work are:
\vspace{-10 pt}
\begin{itemize}
    \item An online learning-based method for generating collision-free, 
    kinodynamically feasible trajectories directly in continuous state and control 
    spaces.
    \item A bidirectional parallel planning architecture that enhances global 
    connectivity while preserving local trajectory quality.
    \item A spatial hashing mechanism within the \texttt{MotionTree} data 
    structure that enables efficient multi-stage feasibility verification.
    \item Extensive experimental validation on ground vehicles with unicycle and 
    bicycle models, and aerial vehicles with quadrotor dynamics.
\end{itemize}

    \section{Related Work}

Sampling-based motion planners (SBMPs) have emerged as dominant paradigms for kinodynamic motion planning in complex environments~\cite{karaman2011sampling}. Sampling-based methods construct feasible trajectories through stochastic exploration of continuous state and control spaces while explicitly respecting dynamic constraints~\cite{kavraki2002probabilistic, lavalle2001randomized, li2016asymptotically}. These algorithms provide probabilistic completeness and strong global exploration capabilities; however, their solutions are often suboptimal and may require post-processing. Optimization-based approaches such as CHOMP, STOMP, and TrajOpt formulate motion planning as nonlinear trajectory optimization~\cite{ratliff2009chomp, kalakrishnan2011stomp, schulman2014motion}. Although they generate smooth, high-quality trajectories, their performance depends heavily on gradient information and good initialization, making them susceptible to local minima in cluttered environments.

To improve exploration efficiency, sampling-based planners have incorporated structural and heuristic refinements. Early methods such as RRT and RRT$^\ast$ rely on uniform random sampling and frequently evaluate edges that do not contribute to the final solution~\cite{lavalle2001randomized, karaman2011sampling}. Subsequent approaches introduced bidirectional search~\cite{kuffner2000rrt}, informed sampling~\cite{gammell2014informed}, and cost-aware edge prioritization~\cite{strub2020adaptively, gammell2015batch} to concentrate computation in promising regions. Batch-informed algorithms such as BIT$^\ast$ integrate graph-search principles with sampling-based planning to reduce redundant computation~\cite{gammell2015batch}. In kinodynamic settings specifically, EST performs forward simulation of randomly sampled controls while biasing expansion toward sparsely explored regions~\cite{hsu2002randomized}, KPIECE leverages low-dimensional projections to guide exploration~\cite{csucan2009kinodynamic}, and SST introduces pruning mechanisms to maintain a sparse set of representative nodes while achieving asymptotic near-optimality~\cite{li2016asymptotically}.

Several approaches adopt a 'plan first, optimize later' paradigm, refining RRT-generated paths using time-optimal steering or bang–bang control primitives~\cite{la2023bang}, an approach supported by empirical evidence that post-optimization can yield high-quality solutions faster than running asymptotically optimal variants such as RRT*~\cite{karaman2011sampling} or SST*~\cite{li2016asymptotically} to convergence~\cite{luo2014empirical}.

Receding-horizon control (RHC) offers an alternative perspective by repeatedly solving short-horizon optimization problems and executing only the first control action before replanning~\cite{Mayne1991ModelPC}. This paradigm has been successfully applied to unified motion planning and control in dynamic environments~\cite{zhang_improved_2019}. Reactive planners such as the Dynamic Window Approach and Hybrid Reciprocal Velocity Obstacles operate over short temporal windows directly in velocity space~\cite{newaz2024adaptive, snape2011hybrid}. However, these approaches generally sacrifice long-horizon planning capability for reactivity.

Bayesian optimization (BO), combined with Gaussian Processes (GPs), provides a principled framework for data-efficient optimization under uncertainty and has recently been explored for motion planning~\cite{quintero2021motion}. GP-based planners model trajectory costs probabilistically and use acquisition functions to guide exploration toward safe and promising regions of the search space~\cite{ungredda2021bayesian}. Extensions such as GPMP and GPMP2 leverage sparse GP representations to enable efficient continuous-time trajectory optimization under kinodynamic constraints~\cite{mukadam2018continuous}. Nevertheless, many BO-based planners remain computationally expensive and are typically applied in unidirectional, locally greedy manners.
Bayesian Optimization over Windows (BOW) addresses this limitation by applying BO over short receding-horizon control windows in continuous state and control spaces. By leveraging GP surrogates to guide control sampling, BOW enables locally optimal and dynamically feasible trajectory generation while explicitly accounting for uncertainty~\cite{raxit2025bow}. Related constrained-BO
planners have also used local cost-map learning for multi-robot kinodynamic
planning under STL specifications~\cite{raxit2026multi}. However, unidirectional window-based BO planners may struggle with global exploration, particularly in environments with narrow passages or complex connectivity structures.

Despite these advances, existing parallel SBMP algorithms focus on RRT variants and do not learn cost maps to efficiently explore the space, limiting their overall performance. To address this gap, we develop BOWConnect, a CPU-based multithreaded planner that leverages high-level parallelism (see definition in~\cite{huang2025prrtc}) of the local BOW planner to surpass state-of-the-art kinodynamic motion planning performance.

    \section{Problem Formulation}

Let $\mathcal{C}$ be the robot configuration space, modeled as an $n$-dimensional smooth 
manifold. Let $\mathcal{C}_{\text{free}} \subset \mathcal{C}$ denote the open subset of 
configurations in which the robot does not intersect any obstacle. The kinodynamic planning 
problem extends beyond geometric path planning by incorporating the robot's dynamics. Let 
$\mathbf{x} = (q, \dot{q}) \in \mathcal{X}$ be the $2n$-dimensional state vector, where 
$\mathcal{X} = T\mathcal{C}$ is the tangent bundle of $\mathcal{C}$. The feasible state 
space is defined as:
\begin{equation}
    \mathcal{X}_{\text{free}} = \{(q, \dot{q}) \in \mathcal{X} \mid q \in 
    \mathcal{C}_{\text{free}}\}
\end{equation}

The system dynamics are governed by the ordinary differential equation 
$\dot{\mathbf{x}} = f(\mathbf{x}, \mathbf{u})$, where $\mathbf{u} \in \mathcal{U} \subset 
\mathbb{R}^m$ is a control input drawn from a compact action set. For a candidate control 
$\mathbf{u}$, the system dynamics are integrated, e.g., using fourth-order Runge-Kutta over a 
finite time horizon $T$ to obtain the predicted trajectory:
\begin{equation}
    \tau(\mathbf{u}) = \Phi(\mathbf{x}, \mathbf{u}) = 
    \{\mathbf{x}(t) : t \in [0, T]\}
\end{equation}

\begin{problem}[Kinodynamic Planning]
Given initial and goal states $\mathbf{x}_I, \mathbf{x}_G \in \mathcal{X}_{\text{free}}$, 
find a control sequence $\mathbf{u} : [0, t_F] \rightarrow \mathcal{U}$ such that the 
resulting trajectory $\tau(\mathbf{u}) = \Phi(\mathbf{x}_I, \mathbf{u})$ satisfies:
\begin{equation}
    \tau(\mathbf{u}) \subset \mathcal{X}_{\text{free}}, \quad 
    \mathbf{x}(t_F) = \mathbf{x}_G
\end{equation}
When time-optimality is required, $t_F$ is minimized over all feasible solutions.
\end{problem}

    \section{BOWConnect Algorithm}

\vspace{10pt}
\begin{algorithm}[t]
\caption{BOWConnect Algorithm}
\label{alg:bowconnect}
\begin{algorithmic}[1]
\REQUIRE Initial state $\mathbf{x}_{\text{start}}$, goal position $\mathbf{p}_{\text{goal}}$, sampling radius $r_{\text{goal}}$, number of workers $N$, time budget $T_{\max}$
\ENSURE Kinodynamic trajectory $\tau$ or failure
\STATE $\mathcal{S} \leftarrow$ \textsc{SampleStates}($\mathbf{x}_{\text{start}}$, $r_{\text{goal}}$, $N$)
\STATE $\mathcal{G} \leftarrow$ \textsc{SampleStates}($\mathbf{p}_{\text{goal}}$, $r_{\text{goal}}$, $N$)
\FOR{$i = 0$ to $N-1$ by step $2$}
    \STATE Launch forward worker $W_f[i]$: $\mathcal{S}[i] \rightarrow \mathcal{G}[i]$
    \STATE Launch backward worker $W_b[i+1]$: $\mathcal{G}[i+1] \rightarrow \mathcal{S}[i+1]$
\ENDFOR
\WHILE{time elapsed $< T_{\max}$}
    \FOR{each forward state $\mathbf{x}_f$ and backward state $\mathbf{x}_b$}
        \IF{$\mathbf{x}_b$ nearby $\mathbf{x}_f$ via spatial hash \AND \textsc{IsKinematicallyFeasible}($\mathbf{x}_f$, $\mathbf{x}_b$)}
            \STATE $(ok, \tau_{\text{bridge}}) \leftarrow$ \textsc{SolveBVP}($\mathbf{x}_f$, $\mathbf{x}_b$)
            \IF{$ok$}
                \STATE \textbf{return} $(true,$ \textsc{MergeTrees}($w_f$, $\tau_{\text{bridge}}$, $w_b$)$)$
            \ENDIF
        \ENDIF
    \ENDFOR
    \IF{any worker reached $\mathbf{p}_{\text{goal}}$}
        \STATE \textbf{return} unidirectional solution
    \ENDIF
\ENDWHILE
\STATE \textbf{return} $(false, \emptyset)$
\end{algorithmic}
\end{algorithm}

The algorithmic steps of BOWConnect
are outlined in Alg.~\ref{alg:bowconnect}.
The core idea is to 
leverage multiple independent BOW planner~\cite{raxit2025bow} instances within a tree-based exploration framework. Unlike traditional bidirectional planners that grow a single forward and backward tree, BOWConnect spawns $N / 2$ forward workers and $N / 2$ backward workers, each maintaining its own search tree and BOW instance. This high-level parallelism enables diverse exploration patterns while preserving the kinodynamic feasibility guarantees of the underlying BOW planner.

Given an initial state $\mathbf{x}_{\text{start}} \in \mathbb{R}^n$ and goal position $\mathbf{p}_{\text{goal}} \in \mathbb{R}^d$, BOWConnect first samples multiple initial configurations around both regions. Specifically, it generates collision-free start states $\mathcal{S} = \{\mathbf{x}_1, \ldots, \mathbf{x}_N\}$ and goal states $\mathcal{G} = \{\mathbf{x}'_1, \ldots, \mathbf{x}'_N\}$ by sampling uniformly within radius $r_{\text{goal}}$ of the respective centers. Each sampled state includes a random heading $\theta \in [-\pi, \pi]$, enabling workers to explore from different initial orientations. This sampling strategy addresses heading ambiguity at the goal and provides diversified exploration directions, significantly improving connection probability between trees.

\subsection{Tree Growth with Constrained Bayesian Optimization}
\begin{figure}[t]
    \centering
    \includegraphics[width=0.8\linewidth]{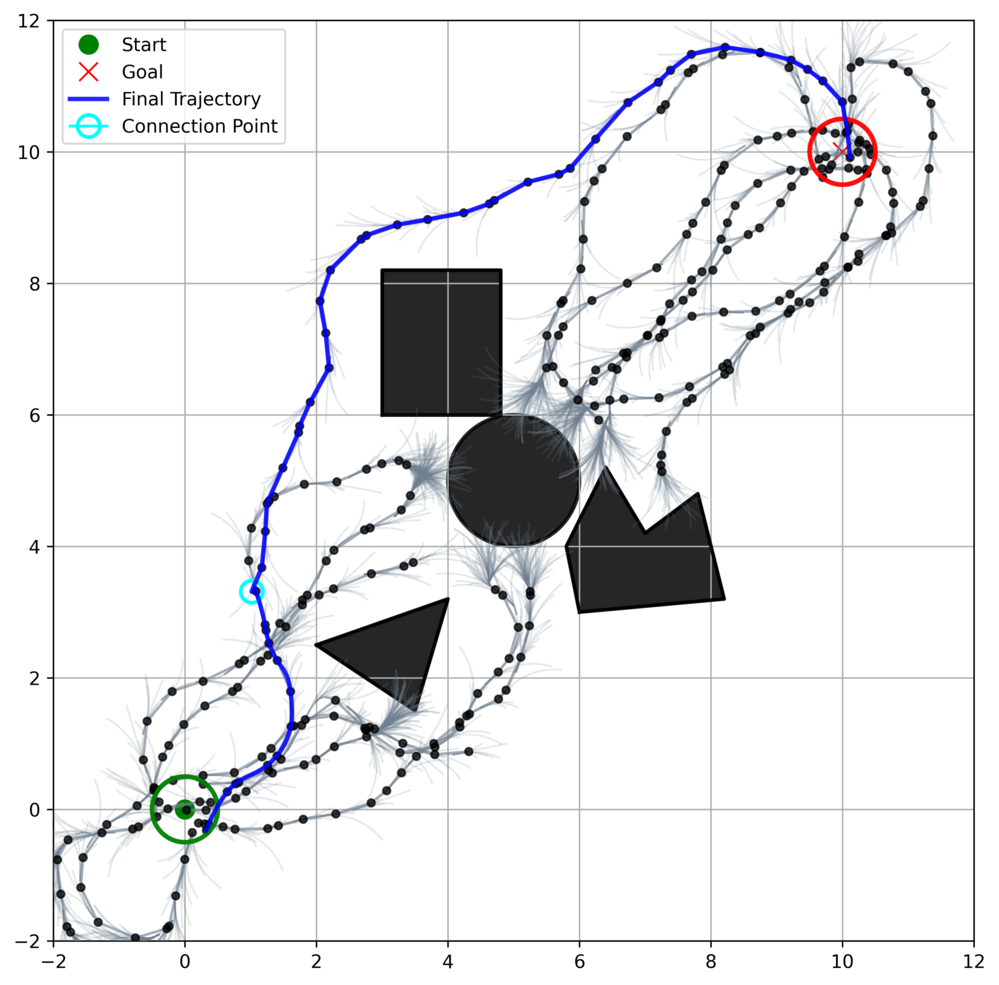}
    \caption{\small{Parallel motion tree growth of \textbf{BOW-Connect} in a cluttered planar environment. Trees are initialized from the start (green) and goal (red) regions and extended in separate threads using the BOW planner~\cite{raxit2025bow}. Grey trajectories depict sampled short-horizon trajectories and the bold blue curve indicates the final connected path. A successful bidirectional connection via boundary value problem solution is highlighted in cyan.}}
    \label{fig:bow_connect_example}
\end{figure}

Figure~\ref{fig:bow_connect_example} illustrates the tree growth behavior of BOW-Connect in a planar environment.
Each worker thread maintains a \texttt{MotionTree} data structure that stores explored states and their parent relationships. The tree growth process follows the RRT paradigm but uses BOW as the steering function instead of simple geometric interpolation. At each iteration, a worker samples a random state $\mathbf{x}_{\text{rand}}$ in the configuration space, finds the nearest state $\mathbf{x}_{\text{near}}$ in its current tree, and invokes BOW to generate a kinodynamically feasible trajectory connecting them.

The BOW planner formulates local planning as a constrained Bayesian Optimization problem over the control space $\mathcal{U} \subset \mathbb{R}^m$, where the dimension $m$ depends on the robot type. For each candidate control $\mathbf{u}$, BOW uses a robot-specific motion model to simulate forward trajectory through numerical integration. The motion model captures the system dynamics via ordinary differential equations $\dot{\mathbf{x}} = f(\mathbf{x}, \mathbf{u})$, which are integrated using fourth-order Runge-Kutta to obtain the predicted trajectory $\tau(\mathbf{u})$ over a finite time horizon $T$. 
For each simulated trajectory, BOW evaluates two functions: a reward and a constraint satisfaction indicator. The reward is defined as the negative distance to the local target for collision-free trajectories:
\begin{equation}
r(\mathbf{u}) = \begin{cases}
-\|\mathbf{p}(T) - \mathbf{p}_{\text{target}}\| & \text{if trajectory is collision-free} \\
-\infty & \text{otherwise}
\end{cases}\label{eqn:reward}
\end{equation}
where $\mathbf{p}(T)$ represents the position component of the predicted state after time horizon $T$. The constraint function returns 1 if the entire trajectory avoids obstacles and 0 otherwise:
\begin{equation}
c(\mathbf{u}) = 1 - \mathbb{I}_{\text{collision}}(\tau(\mathbf{u}))\label{eqn:constraints}
\end{equation}

BOW learns two separate Gaussian Process models: one for the reward $\mathcal{GP}_r$ and one for the constraint $\mathcal{GP}_c$. These models provide mean predictions $\mu_r(\mathbf{u})$, $\mu_c(\mathbf{u})$ and uncertainties $\sigma_r^2(\mathbf{u})$, $\sigma_c^2(\mathbf{u})$ respectively. \textbf{The acquisition function balances exploration and exploitation while respecting collision constraints through the probability of feasibility}:
\begin{equation}
P_{\text{feas}}(\mathbf{u}) = \Phi\left(\frac{\mu_c(\mathbf{u})}{\sigma_c(\mathbf{u})}\right)
\end{equation}
where $\Phi$ is the standard normal cumulative distribution function. The next control to evaluate is selected by optimizing the constrained acquisition function $\alpha(\mathbf{u}) = [\mu_r(\mathbf{u}) - \kappa \sigma_r(\mathbf{u})] \times P_{\text{feas}}(\mathbf{u})$. 

\textbf{A key property of this formulation is its learning behavior in narrow passages. When the reward function in Eqn.~\eqref{eqn:reward} drives the robot toward a region that is largely infeasible, constraint violations inflate $P_{\text{feas}}$ penalties, causing the feasibility term to dominate $\alpha(\mathbf{u})$ and redirecting the search toward collision-free controls in free space.} This mechanism prevents BOW from becoming trapped in local minima induced by inadmissible costs within the planning horizon $T$. When BOW successfully produces a collision-free trajectory $\tau = [\mathbf{x}_{\text{near}}, \mathbf{x}_1, \ldots, \mathbf{x}_m]$, all states are appended sequentially to the tree. Kinodynamic feasibility is guaranteed by construction, as the motion model integration inherently respects velocity, acceleration, and curvature limits. Combined with parallel tree growth across multiple workers, this property gives BOW-Connect space exploration characteristics akin to those of global planners.

\subsection{Connection Detection and BVP Solving}

While workers grow their trees independently, the main thread periodically checks for potential connections between forward and backward trees. To enable efficient connection queries, BOWConnect employs spatial hashing through the \texttt{MotionTree} structure. Each state $\mathbf{x}$ is mapped to a discrete grid cell via the hash function:
\begin{equation}
h(\mathbf{x}) = \lfloor (x - x_0)/\Delta \rfloor \oplus \lfloor (y - y_0)/\Delta \rfloor \oplus \lfloor (z - z_0)/\Delta \rfloor
\end{equation}
where $\Delta$ is the grid resolution and $\oplus$ denotes bit-packing operations. This allows $\mathcal{O}(1)$ average-case lookup to determine if a forward tree state has nearby backward tree states.

When a potential connection is detected, BOWConnect performs a multi-stage verification. First, it checks kinematic feasibility by ensuring the required heading change and travel distance are within vehicle constraints. Specifically, given two states $\mathbf{x}_f$ and $\mathbf{x}_b$ belonging to the forward tree $\mathcal{T}_I$ and backward tree $\mathcal{T}_G$ respectively, the connection is kinematically feasible if:
\begin{equation}
    \left|\theta_{\text{req}} - \theta_f\right| < \theta_{\max} \quad \text{and} \quad \frac{\|\mathbf{p}_b - \mathbf{p}_f\|}{v_{\max}} < T_{\max}
\end{equation}
where $\theta_{\text{req}} = \arctan2(y_b - y_f, x_b - x_f)$ is the required heading and $\mathbf{p}(t)$ denotes the position component of $\mathbf{x}(t)$ at time $t$.

If kinematically feasible, BOWConnect solves a Boundary Value Problem (BVP)~\cite{lavalle2006planning} to generate the connecting trajectory. The BVP seeks a control $\mathbf{u}_c : [0, t_c] \rightarrow \mathcal{U}$ such that the connecting trajectory $\tau(\mathbf{u}_c) = \Phi(\mathbf{x}_f, \mathbf{u}_c)$ satisfies:
\begin{equation}
    \mathbf{x}(t_c) = \mathbf{x}_b, \quad \tau(\mathbf{u}_c) \subset \mathcal{X}_{\text{free}}
\end{equation}
The BVP solver employs proportional control to smoothly transition between states while respecting kinodynamic constraints. Starting from $\mathbf{x}_f$, it iteratively computes the heading error $\Delta\theta = \theta_{\text{desired}} - \theta_{\text{current}}$ and applies yaw rate $\omega = \text{clamp}(k_p \Delta\theta / \Delta t, -\omega_{\max}, \omega_{\max})$ while moving toward $\mathbf{x}_b$ with speed proportional to remaining distance. The resulting trajectory undergoes collision checking before acceptance. If multiple connections are found, BOWConnect selects the one with minimum Euclidean distance, as shorter connections are more reliable and easier to verify.

\subsection{Solution Extraction and Fallback Mechanisms}

BOWConnect terminates when trees successfully connect or when any worker reaches the goal region. Upon connection, the algorithm extracts trajectories from both trees up to the connection points, generates the BVP bridge trajectory, and concatenates them. If the backward trajectory does not fully reach the goal, an additional BVP call extends it from the last state to the goal configuration.

The algorithm implements graceful degradation through fallback solutions. If no connection is found within the time budget but a forward worker reached the goal, that unidirectional solution is returned. Similarly, backward solutions serve as fallbacks. 
\textbf{The parallel architecture provides theoretical speedup proportional to the number of workers, though practical speedup is limited by connection checking overhead and sequential BVP solving}. With $N_f$ forward and $N_b$ backward workers, the probability of finding at least one successful path is:
\begin{equation}
P_{\text{success}} = 1 - (1 - p_{\text{single}})^{N_f \times N_b}
\end{equation}
where $p_{\text{single}}$ is the success probability of a single worker pair. This portfolio effect, combined with BOW's efficient local planning, enables BOWConnect to solve complex kinodynamic planning problems in real-time.

    \section{Experiments and Results}
All experiments were performed on a desktop computer running Ubuntu 22.04 LTS with an AMD Ryzen\textsuperscript{TM} 9 7950X processor with 32\,GB of RAM. Robot localization was achieved using a VICON motion capture system sampling at 120\,Hz. The BOWConnect Planner was parameterized with the following constraints: maximum and minimum speeds of 1.0\,m/s and 0.0\,m/s respectively, maximum acceleration of 0.5\,m/s\textsuperscript{2}, maximum yaw rate of 0.6981\,rad/s, and maximum yaw acceleration of 2.0472\,rad/s\textsuperscript{2}. The planner employed a time discretization of 0.1\,s and a local planning horizon of 3.0 - 5.0\,s. 
We evaluate \textit{BOWConnect} against state-of-the-art kinodynamic planners with OMPL implementation including RRT~\cite{lavalle2001randomized}, SST~\cite{li2016asymptotically}, EST~\cite{hsu2002randomized}, KPIECE~\cite{csucan2009kinodynamic}, and BOW~\cite{raxit2025bow}.

\renewcommand{\arraystretch}{0.8}

\newcommand{\model}[1]{\makecell[l]{\scriptsize #1}}

\begin{table*}[!t]
\centering

\begin{adjustbox}{max width=0.9\textwidth,}
\begin{tabular}{|c|l|l|r|r|r|r|r|}
\toprule
\textbf{Environment} & \textbf{Planner} & \textbf{Model} &
\textbf{Total Time (s)} & \textbf{Traj. Length} & \textbf{Succ. (\%)} & \textbf{Avg. Velocity} & \textbf{Avg. Jerk} \\
\midrule

\multirow{12}{*}{%
  \begin{tikzpicture}
    \node[inner sep=0] (img) {\includegraphics[width=0.22\textwidth, trim=80 2 85 2,
  clip]{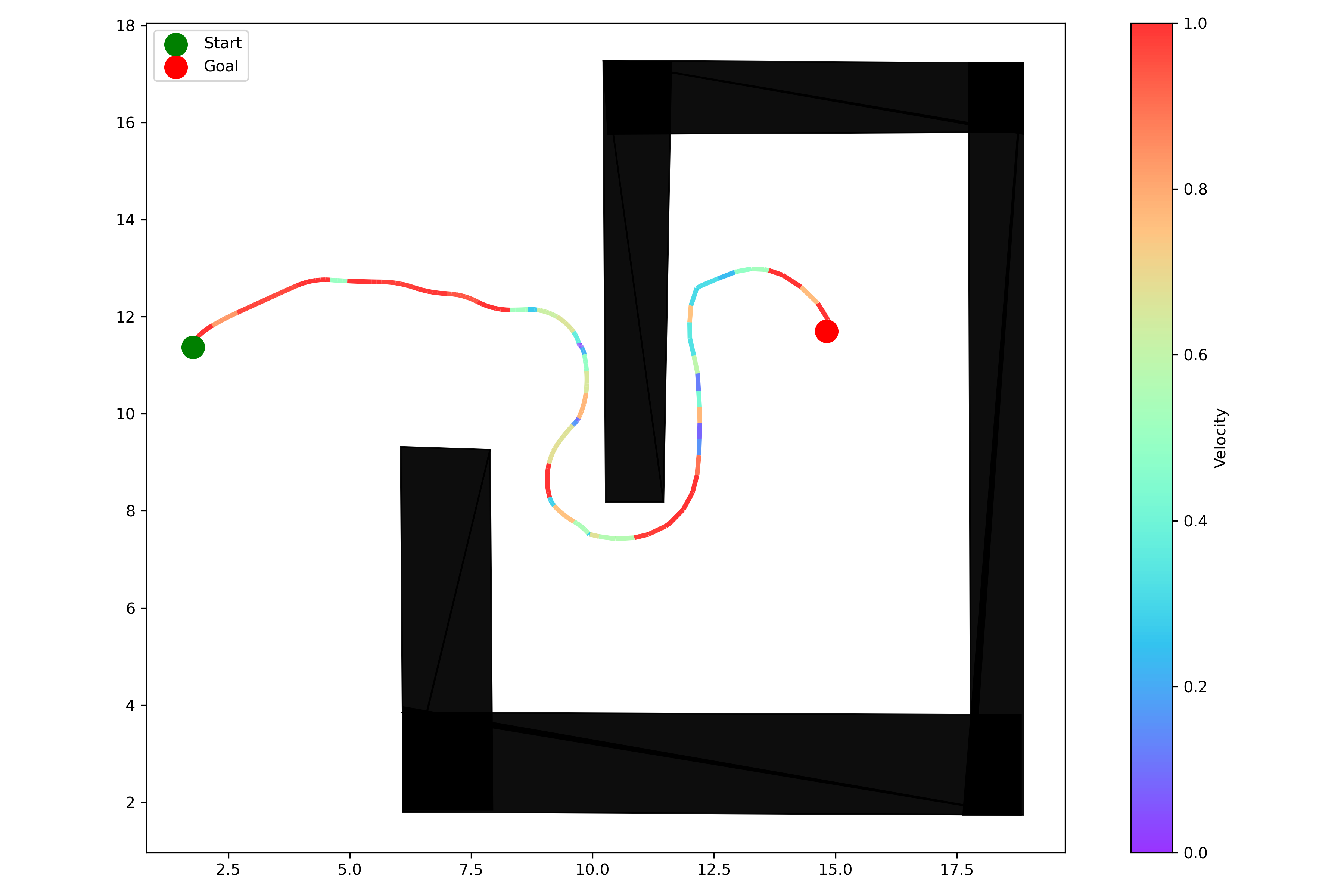}};
    \node[fill=white, opacity=0.85, text opacity=1, rounded corners, inner sep=2pt]
      at ([yshift=-3.3pt]img.south) {\textbf{Bugtrap}};
  \end{tikzpicture}
}
&   \multirow{2}{*}{RRT\cite{lavalle2001randomized}} & \model{Unicycle} & 0.815 $\pm$ 0.734 & 22.648 $\pm$ 3.679 & \textbf{100.0} & 0.781 $\pm$ 0.005 & 0.8871 $\pm$ 0.0698 \\
\cline{3-8}
&    & \model{Bicycle} & \textemdash & \textemdash & 0.0 & \textemdash & \textemdash \\
\cline{2-8}
&   \multirow{2}{*}{SST\cite{li2016asymptotically}} & \model{Unicycle} & 5.066 $\pm$ 0.007 & 24.601 $\pm$ 3.493 & \textbf{100.0} & 0.794 $\pm$ 0.024 & 0.8406 $\pm$ 0.0494 \\
\cline{3-8}
&    & \model{Bicycle} & \textemdash & \textemdash & 0.0 & \textemdash & \textemdash \\
\cline{2-8}
&   \multirow{2}{*}{EST\cite{hsu2002randomized}} & \model{Unicycle} & 0.849 $\pm$ 0.569 & \textbf{20.937 $\pm$ 1.869} & \textbf{100.0} & 0.808 $\pm$ 0.037 & 0.8982 $\pm$ 0.0806 \\
\cline{3-8}
&    & \model{Bicycle} & \textemdash & \textemdash & 0.0 & \textemdash & \textemdash \\
\cline{2-8}
&   \multirow{2}{*}{KPIECE\cite{csucan2009kinodynamic}} & \model{Unicycle} & 2.220 $\pm$ 2.591 & 22.752 $\pm$ 4.194 & 60.0 & \textbf{0.851 $\pm$ 0.016} & 0.7024 $\pm$ 0.0464 \\
\cline{3-8}
&    & \model{Bicycle} & \textemdash & \textemdash & 0.0 & \textemdash & \textemdash \\
\cline{2-8}
&   \multirow{2}{*}{BOW\cite{raxit2025bow}} & \model{Unicycle} & 0.200 $\pm$ 0.156 & 45.324 $\pm$ 35.758 & \textbf{100.0} & 0.735 $\pm$ 0.052 & \textbf{0.1373 $\pm$ 0.0102} \\
\cline{3-8}
&    & \model{Bicycle} & \textbf{0.019 $\pm$ 0.006} & 46.536 $\pm$ 10.557 & \textbf{100.0} & \textbf{1.060 $\pm$ 0.033} & \textbf{0.0065 $\pm$ 0.0010} \\
\cline{2-8}
&   \multirow{2}{*}{BOWConnect} & \model{Unicycle} & \textbf{0.035 $\pm$ 0.014} & 22.091 $\pm$ 3.427 & \textbf{100.0} & 0.800 $\pm$ 0.045 & 0.1881 $\pm$ 0.0289\\
\cline{3-8}
&    & \model{Bicycle} & 0.021 $\pm$ 0.005 & \textbf{23.838 $\pm$ 2.397} & \textbf{100.0} & 1.000 $\pm$ 0.006 & 0.0279 $\pm$ 0.0047 \\

\midrule

\multirow{12}{*}{%
  \begin{tikzpicture}
    \node[inner sep=0] (img) {\includegraphics[width=0.22\textwidth, trim=115 2 85 2,
  clip]{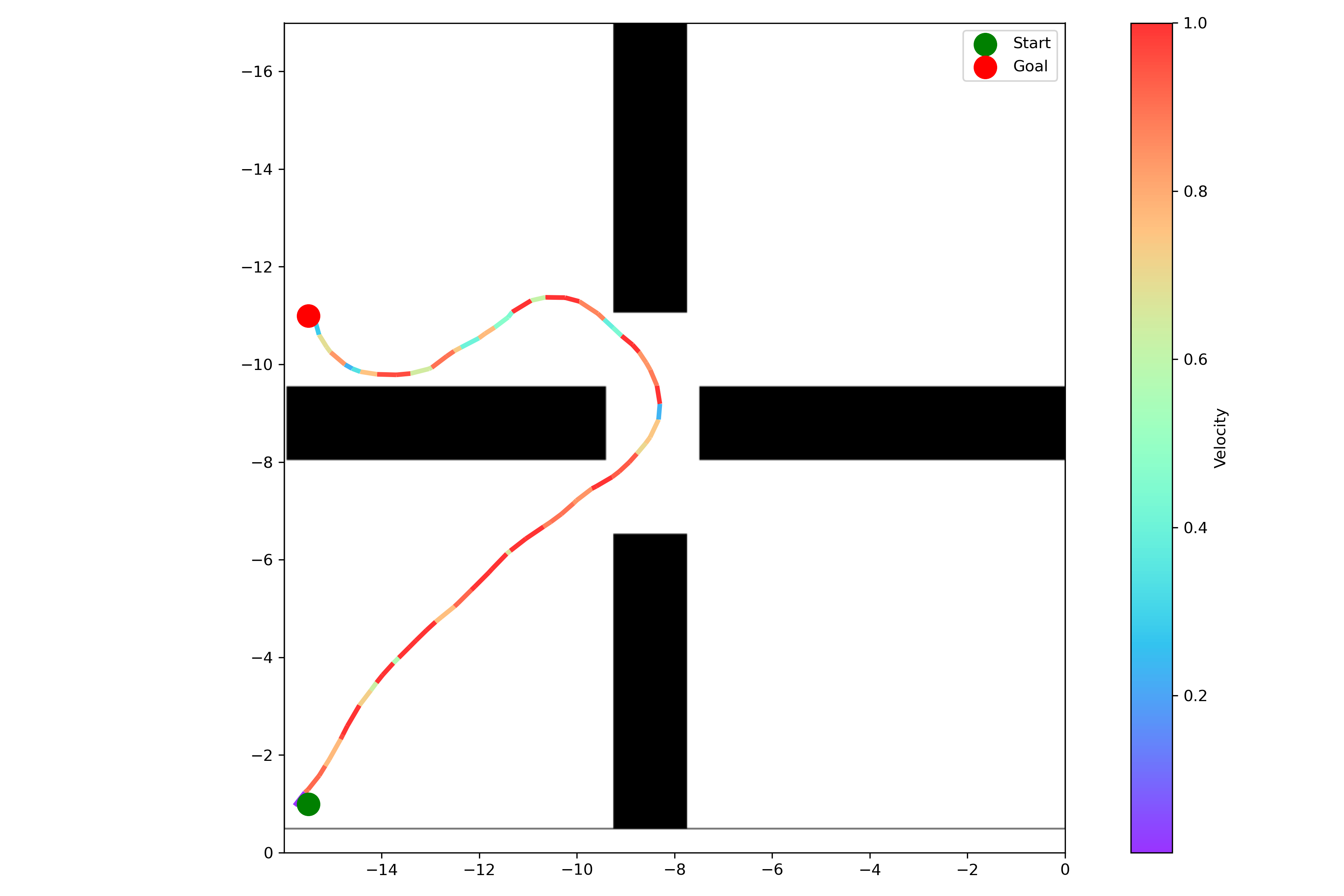}};
    \node[fill=white, opacity=0.85, text opacity=1, rounded corners, inner sep=2pt]
      at ([yshift=-3pt]img.south) {\textbf{Narrow Passage-1}};
  \end{tikzpicture}
}
&   \multirow{2}{*}{RRT\cite{lavalle2001randomized}} & \model{Unicycle} & 0.903 $\pm$ 0.606 & 21.714 $\pm$ 1.363 & \textbf{100.0} & 0.732 $\pm$ 0.05 & 0.8528 $\pm$ 0.0320 \\
\cline{3-8}
&    & \model{Bicycle} & 25.989 $\pm$ 4.714 & 74.487 $\pm$ 6.992 & 80.0 & 0.765 $\pm$ 0.08 & 0.7982 $\pm$ 0.0474 \\
\cline{2-8}
&   \multirow{2}{*}{SST\cite{li2016asymptotically}} & \model{Unicycle} & 30.214 $\pm$ 0.009 & 20.789 $\pm$ 1.135 & \textbf{100.0} & 0.794 $\pm$ 0.007 & 0.8618 $\pm$ 0.0414 \\
\cline{3-8}
&    & \model{Bicycle} & 30.091 $\pm$ 0.009 & 74.640 $\pm$ 8.995 & 80.0 & 0.784 $\pm$ 0.032 & 0.7549 $\pm$ 0.0588 \\
\cline{2-8}
&   \multirow{2}{*}{EST\cite{hsu2002randomized}} & \model{Unicycle} & 6.112 $\pm$ 7.231 & \textbf{20.090 $\pm$ 0.537} & \textbf{100.0} & 0.787 $\pm$ 0.029 & 0.8813 $\pm$ 0.0520 \\
\cline{3-8}
&    & \model{Bicycle} & \textemdash & \textemdash & 0.0 & \textemdash & \textemdash \\
\cline{2-8}
&   \multirow{2}{*}{KPIECE\cite{csucan2009kinodynamic}} & \model{Unicycle} & 18.932 $\pm$ 11.173 & 20.853 $\pm$ 0.889 & \textbf{100.0} & \textbf{0.813 $\pm$ 0.030} & 0.7052 $\pm$ 0.0201 \\
\cline{3-8}
&    & \model{Bicycle} & \textemdash & \textemdash & 0.0 & \textemdash & \textemdash \\
\cline{2-8}
&   \multirow{2}{*}{BOW\cite{raxit2025bow}} & \model{Unicycle} & 19.538 $\pm$ 13.017 & 53.613 $\pm$ 20.695 & \textbf{100.0} & 0.610 $\pm$ 0.041 & \textbf{0.1677 $\pm$ 0.0047} \\
\cline{3-8}
&    & \model{Bicycle} & \textemdash & \textemdash & 0.0 & \textemdash & \textemdash \\
\cline{2-8}
&   \multirow{2}{*}{BOWConnect} & \model{Unicycle} & \textbf{0.048 $\pm$ 0.003} & 24.801 $\pm$ 6.550 & \textbf{100.0} & 0.783 $\pm$ 0.077 & 0.3740 $\pm$ 0.1797 \\
\cline{3-8}
&    & \model{Bicycle} & \textbf{0.034 $\pm$ 0.008} & \textbf{22.190 $\pm$ 2.059} & \textbf{100.0} & \textbf{0.992 $\pm$ 0.040} & \textbf{0.0381 $\pm$ 0.0064} \\

\midrule

\multirow{12}{*}{%
  \begin{tikzpicture}
    \node[inner sep=0] (img) {\includegraphics[width=0.17\textwidth, trim=10 0 40 0,
  clip]{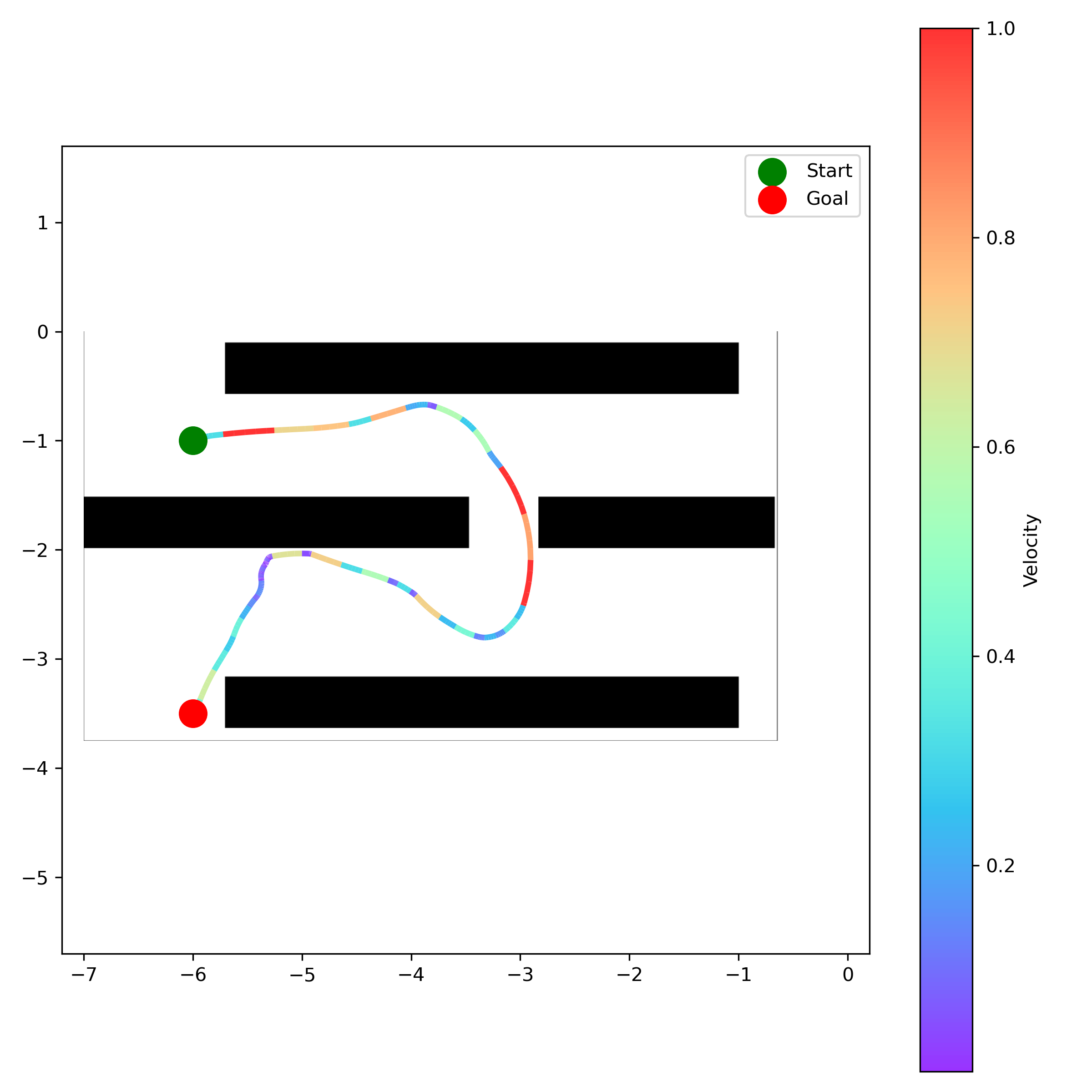}};
    \node[fill=white, opacity=0.85, text opacity=1, rounded corners, inner sep=2pt]
      at ([yshift=-2pt]img.south) {\textbf{Narrow Passage-2}};
  \end{tikzpicture}
}
&   \multirow{2}{*}{RRT\cite{lavalle2001randomized}} & \model{Unicycle} & 0.357 $\pm$ 0.307 & 11.374 $\pm$ 3.418 & \textbf{100.0} & \textbf{0.710 $\pm$ 0.083} & 0.7586 $\pm$ 0.0364 \\
\cline{3-8}
&    & \model{Bicycle} & 26.528 $\pm$ 6.883 & \textbf{8.229 $\pm$ 10.622} & 40.0 & 0.477 $\pm$ 0.244 & 0.5550 $\pm$ 0.2486 \\
\cline{2-8}
&   \multirow{2}{*}{SST\cite{li2016asymptotically}} & \model{Unicycle} & 30.062 $\pm$ 0.009 & \textbf{6.596 $\pm$ 0.165} & \textbf{100.0} & 0.701 $\pm$ 0.028 & 0.7912 $\pm$ 0.1364 \\
\cline{3-8}
&    & \model{Bicycle} & 30.003 $\pm$ 0.001 & 4.252 $\pm$ 8.476 & 20.0 & 0.459 $\pm$ 0.234 & 0.1403 $\pm$ 0.3137 \\
\cline{2-8}
&   \multirow{2}{*}{EST\cite{hsu2002randomized}} & \model{Unicycle} & 0.384 $\pm$ 0.206 & 8.434 $\pm$ 1.392 & \textbf{100.0} & 0.676 $\pm$ 0.056 & 0.8347 $\pm$ 0.0642 \\
\cline{3-8}
&    & \model{Bicycle} & \textemdash & \textemdash & 0.0 & \textemdash & \textemdash \\
\cline{2-8}
&   \multirow{2}{*}{KPIECE\cite{csucan2009kinodynamic}} & \model{Unicycle} & 1.432 $\pm$ 0.999 & 8.522 $\pm$ 1.026 & \textbf{100.0} & 0.700 $\pm$ 0.068 & 0.6730 $\pm$ 0.0969 \\
\cline{3-8}
&    & \model{Bicycle} & \textemdash & \textemdash & 0.0 & \textemdash & \textemdash \\
\cline{2-8}
&   \multirow{2}{*}{BOW\cite{raxit2025bow}} & \model{Unicycle} & 6.971 $\pm$ 11.596 & 31.626 $\pm$ 21.971 & \textbf{100.0} & 0.431 $\pm$ 0.036 & \textbf{0.1504 $\pm$ 0.0038} \\
\cline{3-8}
&    & \model{Bicycle} & \textemdash & \textemdash & 0.0 & \textemdash & \textemdash \\
\cline{2-8}
&   \multirow{2}{*}{BOWConnect} & \model{Unicycle} & \textbf{0.056 $\pm$ 0.021} & 8.401 $\pm$ 0.737 & \textbf{100.0} & 0.522 $\pm$ 0.032 & 0.2232 $\pm$ 0.0203 \\
\cline{3-8}
&    & \model{Bicycle} & \textbf{0.048 $\pm$ 0.011} & 8.603 $\pm$ 0.034 & \textbf{100.0} & \textbf{0.641 $\pm$ 0.031} & \textbf{0.0493 $\pm$ 0.0041} \\

\midrule

\multirow{12}{*}{%
  \begin{tikzpicture}
    \node[inner sep=0] (img) {\includegraphics[width=0.21\textwidth, trim=115 2 85 2,
  clip]{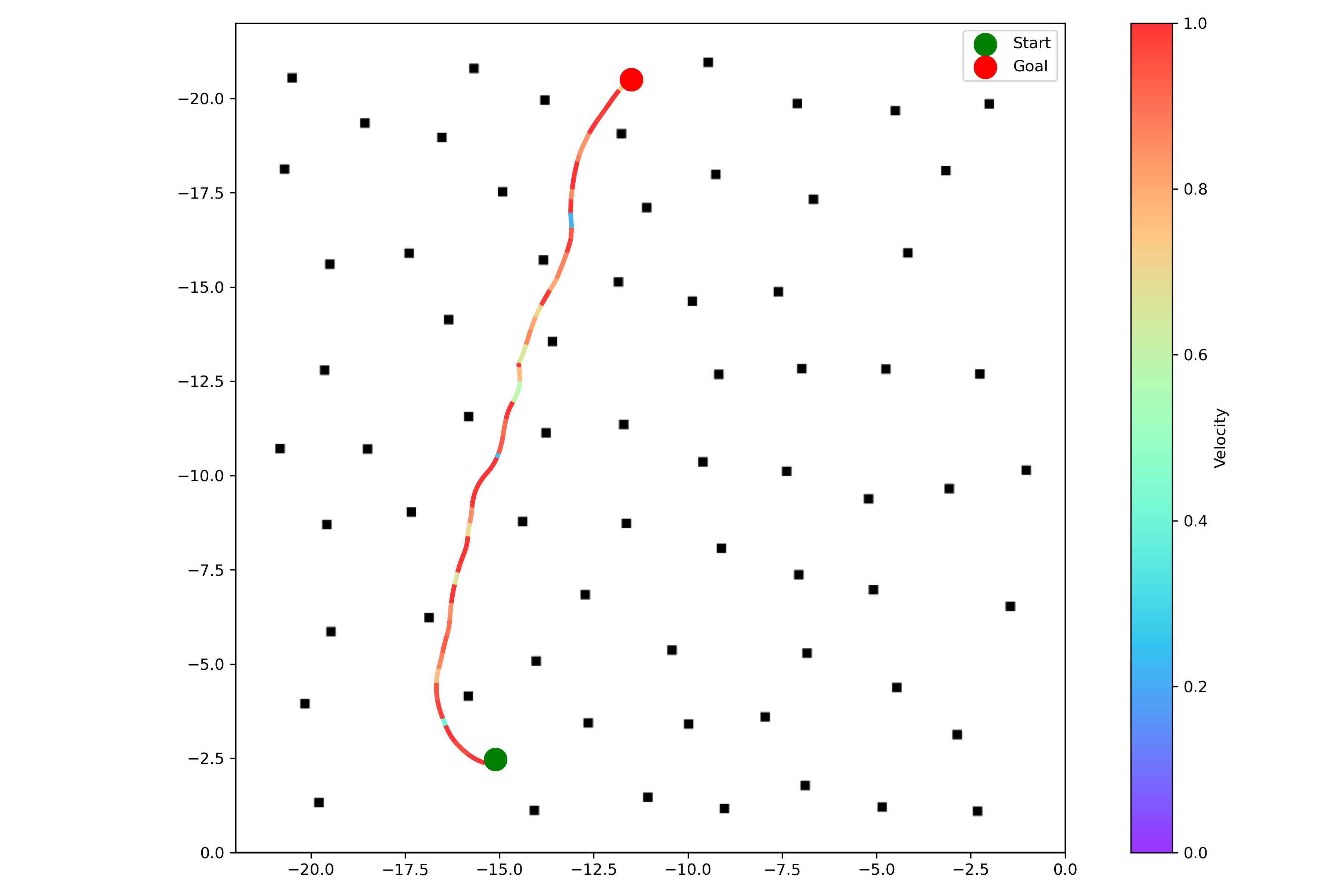}};
    \node[fill=white, opacity=0.85, text opacity=1, rounded corners, inner sep=2pt]
      at ([yshift=-3.3pt]img.south) {\textbf{Forest}};
  \end{tikzpicture}
}
&   \multirow{2}{*}{RRT\cite{lavalle2001randomized}} & \model{Unicycle} & 1.020 $\pm$ 1.104 & 21.954 $\pm$ 1.646 & \textbf{100.0} & 0.782 $\pm$ 0.035 & 0.8671 $\pm$ 0.0249 \\
\cline{3-8}
&    & \model{Bicycle} & 1.148 $\pm$ 0.582 & 62.091 $\pm$ 14.824 & \textbf{100.0} & 0.706 $\pm$ 0.016 & 0.8065 $\pm$ 0.0440 \\
\cline{2-8}
&   \multirow{2}{*}{SST\cite{li2016asymptotically}} & \model{Unicycle} & 5.052 $\pm$ 0.005 & 26.491 $\pm$ 7.449 & \textbf{100.0} & 0.801 $\pm$ 0.012 & 0.8692 $\pm$ 0.0545 \\
\cline{3-8}
&    & \model{Bicycle} & 5.028 $\pm$ 0.002 & 63.091 $\pm$ 3.387 & 60.0 & 0.791 $\pm$ 0.018 & 0.6921 $\pm$ 0.0110 \\
\cline{2-8}
&   \multirow{2}{*}{EST\cite{hsu2002randomized}} & \model{Unicycle} & 2.072 $\pm$ 2.196 & 23.727 $\pm$ 5.727 & 80.0 & 0.807 $\pm$ 0.032 & 0.8942 $\pm$ 0.0426 \\
\cline{3-8}
&    & \model{Bicycle} & \textemdash & \textemdash & 0.0 & \textemdash & \textemdash \\
\cline{2-8}
&   \multirow{2}{*}{KPIECE\cite{csucan2009kinodynamic}} & \model{Unicycle} & 4.309 $\pm$ 1.621 & 30.613 $\pm$ 10.465 & 20.0 &  \textbf{0.817 $\pm$ 0.041} & 0.7486 $\pm$ 0.0601 \\
\cline{3-8}
&    & \model{Bicycle} & \textemdash & \textemdash & 0.0 & \textemdash & \textemdash \\
\cline{2-8}
&   \multirow{2}{*}{BOW\cite{raxit2025bow}} & \model{Unicycle} & 0.045 $\pm$ 0.029 & 24.523 $\pm$ 1.402 & \textbf{100.0} & 0.790 $\pm$ 0.054 & \textbf{0.1671 $\pm$ 0.0126} \\
\cline{3-8}
&    & \model{Bicycle} & 0.030 $\pm$ 0.005 & 29.437 $\pm$ 6.909 & \textbf{100.0} & \textbf{0.971 $\pm$ 0.021} & \textbf{0.0101 $\pm$ 0.0013} \\
\cline{2-8}
&   \multirow{2}{*}{BOWConnect} & \model{Unicycle} & \textbf{0.024 $\pm$ 0.001} & \textbf{20.176 $\pm$ 1.496} & \textbf{100.0} & 0.811 $\pm$ 0.036 & 0.2423 $\pm$ 0.0434 \\
\cline{3-8}
&    & \model{Bicycle} & \textbf{0.025 $\pm$ 0.001} & \textbf{21.464 $\pm$ 2.795} & \textbf{100.0} & 0.871 $\pm$ 0.052 & 0.0193 $\pm$ 0.0075 \\

\midrule

\multirow{12}{*}{%
  \begin{tikzpicture}
    \node[inner sep=0] (img) {\includegraphics[width=0.22\textwidth, trim=115 2 85 2,
  clip]{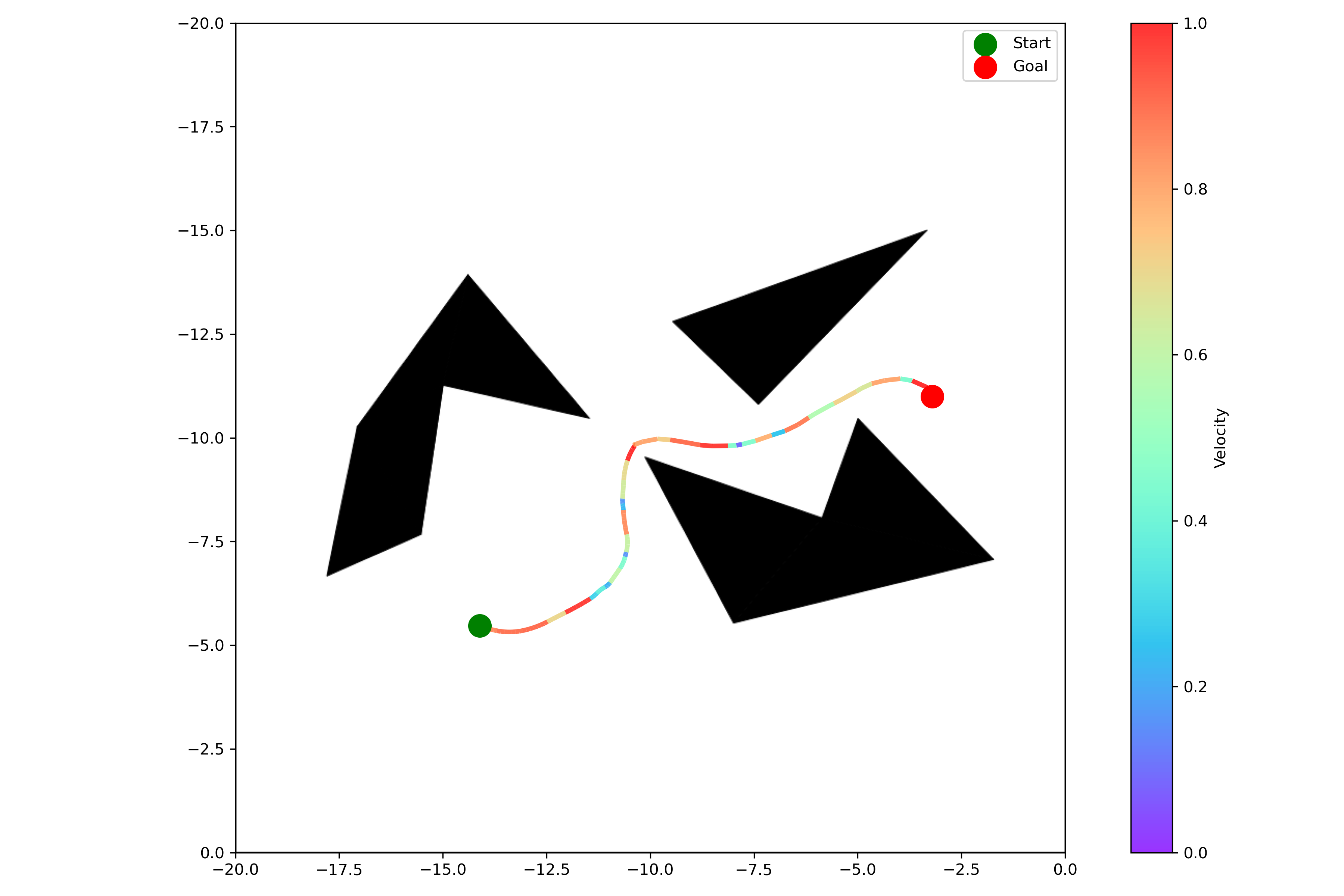}};
    \node[fill=white, opacity=0.85, text opacity=1, rounded corners, inner sep=2pt]
      at ([yshift=-3pt]img.south) {\textbf{Nonconvex}};
  \end{tikzpicture}
}
&   \multirow{2}{*}{RRT\cite{lavalle2001randomized}} & \model{Unicycle} & 0.612 $\pm$ 0.742 & 26.355 $\pm$ 11.862 & \textbf{100.0} & 0.754 $\pm$ 0.033 & 0.8427 $\pm$ 0.0416 \\
\cline{3-8}
&    & \model{Bicycle} & 3.976 $\pm$ 1.480 & 31.496 $\pm$ 12.258 & 40.0 & 0.667 $\pm$ 0.055 & 0.7312 $\pm$ 0.1038 \\
\cline{2-8}
&   \multirow{2}{*}{SST\cite{li2016asymptotically}} & \model{Unicycle} & 5.068 $\pm$ 0.006 & 21.715 $\pm$ 12.541 & 80.0 & 0.764 $\pm$ 0.013 & 0.7782 $\pm$ 0.0797 \\
\cline{3-8}
&    & \model{Bicycle} & 5.021 $\pm$ 0.005 & 35.528 $\pm$ 16.913 & 20.0 & 0.732 $\pm$ 0.057 & 0.6652 $\pm$ 0.1004 \\
\cline{2-8}
&   \multirow{2}{*}{EST\cite{hsu2002randomized}} & \model{Unicycle} & 1.369 $\pm$ 2.241 & 19.414 $\pm$ 8.204 & 80.0 & 0.795 $\pm$ 0.026 & 0.8891 $\pm$ 0.0434 \\
\cline{3-8}
&    & \model{Bicycle} & \textemdash & \textemdash & 0.0 & \textemdash & \textemdash \\
\cline{2-8}
&   \multirow{2}{*}{KPIECE\cite{csucan2009kinodynamic}} & \model{Unicycle} & \textemdash & \textemdash & 0.0 & \textemdash & \textemdash \\
\cline{3-8}
&    & \model{Bicycle} & \textemdash & \textemdash & 0.0 & \textemdash & \textemdash \\
\cline{2-8}
&   \multirow{2}{*}{BOW\cite{raxit2025bow}} & \model{Unicycle} & 0.034 $\pm$ 0.015 & 16.770 $\pm$ 1.044 & \textbf{100.0} & 0.770 $\pm$ 0.079 & \textbf{0.1400 $\pm$ 0.0070} \\
\cline{3-8}
&    & \model{Bicycle} & 0.022 $\pm$ 0.012 & 28.626 $\pm$ 18.668 & 80.0 & 0.849 $\pm$ 0.476 & \textbf{0.0049 $\pm$ 0.0028} \\
\cline{2-8}
&   \multirow{2}{*}{BOWConnect} & \model{Unicycle} & \textbf{0.024 $\pm$ 0.007} & \textbf{15.814 $\pm$ 0.824} & \textbf{100.0} & \textbf{0.859 $\pm$ 0.060} & 0.2187 $\pm$ 0.0397 \\
\cline{3-8}
&    & \model{Bicycle} & \textbf{0.019 $\pm$ 0.004} & \textbf{17.816 $\pm$ 3.969} & \textbf{100.0} & \textbf{0.973 $\pm$ 0.039} & 0.0301 $\pm$ 0.0066 \\

\midrule

\multirow{12}{*}{%
  \begin{tikzpicture}
    \node[inner sep=0] (img) {\includegraphics[width=0.22\textwidth, trim=60 2 43 2,
  clip]{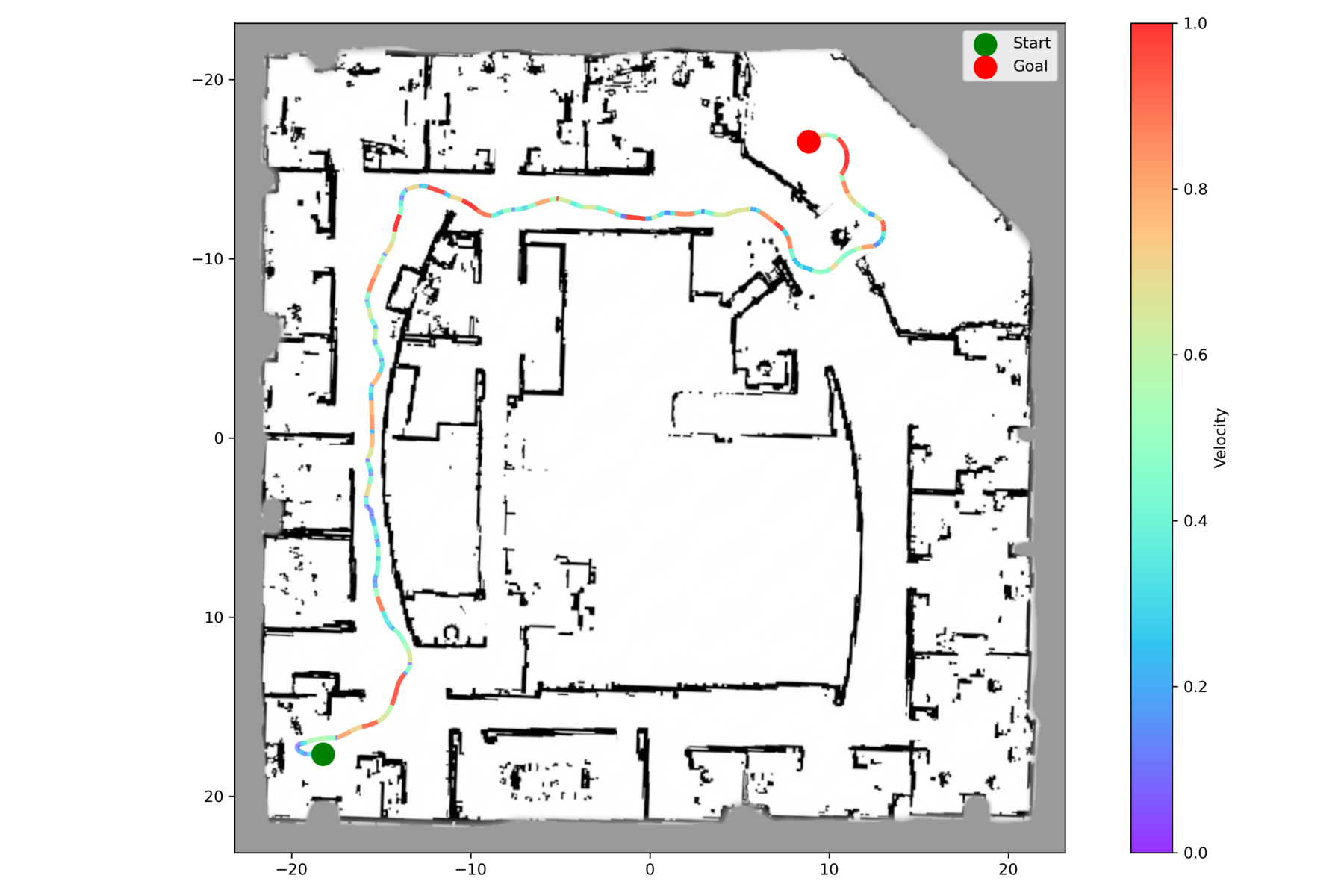}};
    \node[fill=white, opacity=0.85, text opacity=1, rounded corners, inner sep=2pt]
      at ([yshift=-3.0pt]img.south) {\textbf{Intel}};
  \end{tikzpicture}
}
&   \multirow{2}{*}{RRT\cite{lavalle2001randomized}} & \model{Unicycle} & 6.886 $\pm$ 5.033 & 77.852 $\pm$ 4.577 & \textbf{100.0} & 0.777 $\pm$ 0.016 & 0.8563 $\pm$ 0.0153 \\
\cline{3-8}
&    & \model{Bicycle} & 15.945 $\pm$ 3.522 & 164.459 $\pm$ 4.196 & \textbf{100.0} & 0.769 $\pm$ 0.013 & 0.9034 $\pm$ 0.0238 \\
\cline{2-8}
&   \multirow{2}{*}{SST\cite{li2016asymptotically}} & \model{Unicycle} & 30.184 $\pm$ 0.016 & 75.559 $\pm$ 7.137 & 60.0 & 0.781 $\pm$ 0.011 & 0.8389 $\pm$ 0.0303 \\
\cline{3-8}
&    & \model{Bicycle} & 30.088 $\pm$ 0.004 & \textbf{152.552 $\pm$ 11.493} & 60.0 & 0.784 $\pm$ 0.005 & 0.8790 $\pm$ 0.0218 \\
\cline{2-8}
&   \multirow{2}{*}{EST\cite{hsu2002randomized}} & \model{Unicycle} & 23.704 $\pm$ 12.025 & \textbf{70.698 $\pm$ 4.662} & 40.0 & \textbf{0.808 $\pm$ 0.011} & 0.8625 $\pm$ 0.0471 \\
\cline{3-8}
&    & \model{Bicycle} & \textemdash & \textemdash & 0.0 & \textemdash & \textemdash \\
\cline{2-8}
&   \multirow{2}{*}{KPIECE\cite{csucan2009kinodynamic}} & \model{Unicycle} & \textemdash & \textemdash & 0.0 & \textemdash & \textemdash \\
\cline{3-8}
&    & \model{Bicycle} & \textemdash & \textemdash & 0.0 & \textemdash & \textemdash \\
\cline{2-8}
&   \multirow{2}{*}{BOW\cite{raxit2025bow}} & \model{Unicycle} & 12.893 $\pm$ 14.040 & 180.675 $\pm$ 119.789 & 80.0 & 0.430 $\pm$ 0.242 & \textbf{0.1369 $\pm$ 0.0767} \\
\cline{3-8}
&    & \model{Bicycle} & \textemdash & \textemdash & 0.0 & \textemdash & \textemdash \\
\cline{2-8}
&   \multirow{2}{*}{BOWConnect} & \model{Unicycle} & \textbf{0.397 $\pm$ 0.039} & 83.803 $\pm$ 14.019 & \textbf{100.0} & 0.566 $\pm$ 0.009 & 0.1904 $\pm$ 0.0137 \\
\cline{3-8}
&    & \model{Bicycle} & \textbf{2.133 $\pm$ 1.707} & 917.917 $\pm$ 773.043 & \textbf{100.0} & \textbf{1.089 $\pm$ 0.008} & \textbf{0.0118 $\pm$ 0.0023} \\

\bottomrule
\end{tabular}
\end{adjustbox}
\caption{
Quantitative comparison of state-of-the-art kinodynamic motion planners across six environments for a UGV platform under Unicycle and Bicycle motion models, with best results per environment in \textbf{bold} and dashes (\textemdash) indicating failure to find a feasible solution. Overlaid on each environment are qualitative BOWConnect trajectories, where color encodes the linear velocity profile.}
\vspace{-15pt}
\label{tab:ugv_results}
\end{table*}

\subsection{UGV Benchmark Results}
\label{ugv_benchmark}

All planners are evaluated across six complex environments: \textit{Bugtrap}, \textit{Narrow Passage-1}, \textit{Narrow Passage-2}, \textit{Forest}, \textit{Nonconvex}, and \textit{Intel} as shown in Table.~\ref{tab:ugv_results}. For each environment–planner pair, five independent trials were conducted, and the mean and standard deviation of all key metrics (except success rate) were computed. All planners were evaluated under both \textit{Unicycle} and \textit{Bicycle} motion models. The evaluation metrics include total planning time, trajectory length, success rate, average velocity, and average jerk.

We set the solver timeout to $5$--$30$~s depending on environment complexity. Across all environments, \textit{BOWConnect} achieves 100\% success under both Unicycle and Bicycle motion models. Baseline planners degrade significantly under the Bicycle model, which imposes stricter nonholonomic constraints. EST and KPIECE fail entirely (0\%) under the Bicycle model in every environment. RRT and SST also degrade, dropping to 40\% and 20\% in \textit{Narrow Passage-2} and to 40\% and 20\% in \textit{Nonconvex}, respectively. Even under the Unicycle model, KPIECE fails in \textit{Nonconvex} and \textit{Intel}, while SST and EST drop below 100\% in several environments.
\textit{BOWConnect} also achieves the fastest or near-fastest computation time across all environments under both motion models. In \textit{Bugtrap}, it computes solutions in $0.035$~s (Unicycle) and $0.021$~s (Bicycle), outperforming all baselines except BOW under the Bicycle model ($0.019$~s). In \textit{Narrow Passage-1} and \textit{Narrow Passage-2}, where classical planners approach the $30$~s time limit or fail, \textit{BOWConnect} solves both in under $0.06$~s. In \textit{Forest} and \textit{Nonconvex}, it computes solutions in under $0.03$~s under both models. Even in the most complex environment (\textit{Intel}), \textit{BOWConnect} remains the fastest planner at $0.397$~s (Unicycle) and $2.133$~s (Bicycle), while baselines require $6.886$ to $30$~s.

\textit{BOWConnect} also produces shorter trajectories in several environments. In \textit{Nonconvex}, forward-only planners tend to explore suboptimal paths, whereas \textit{BOWConnect}'s backward tree guides expansion toward the goal more directly, achieving the shortest average trajectory ($15.814$~m Unicycle, $17.816$~m Bicycle) among all planners. The benefit is particularly evident under the Bicycle model, where BOW produces $28.626$~m compared to \textit{BOWConnect}'s $17.816$~m. Similarly, in \textit{Forest}, \textit{BOWConnect} achieves the shortest trajectory under both models ($20.176$~m and $21.464$~m).

The average velocity of \textit{BOWConnect} remains competitive across all environments and achieves the highest value in \textit{Nonconvex} for both motion models.
BOW-based planners further exhibit significantly lower jerk values than classical sampling-based methods, indicating smoother control profiles. For example, in \textit{Bugtrap} (Bicycle), \textit{BOWConnect} achieves an average jerk of 0.0279, substantially lower than RRT and EST.

\renewcommand{\arraystretch}{1.1}

\begin{table*}[t!]
\centering

\begin{adjustbox}{max width=\textwidth}
\begin{tabular}{|c|l|r|r|r|r|r|}
\toprule
\textbf{Environment} & \textbf{Planner} &
\textbf{Total Time (s)} & \textbf{Traj. Length} & \textbf{Succ. (\%)} & \textbf{Avg. Velocity} & \textbf{Avg. Jerk} \\
\midrule

\multirow{6}{*}{%
  \begin{tikzpicture}
    \node[inner sep=0] (img) {\includegraphics[width=0.145\textwidth, trim=120 100 85 140, clip]{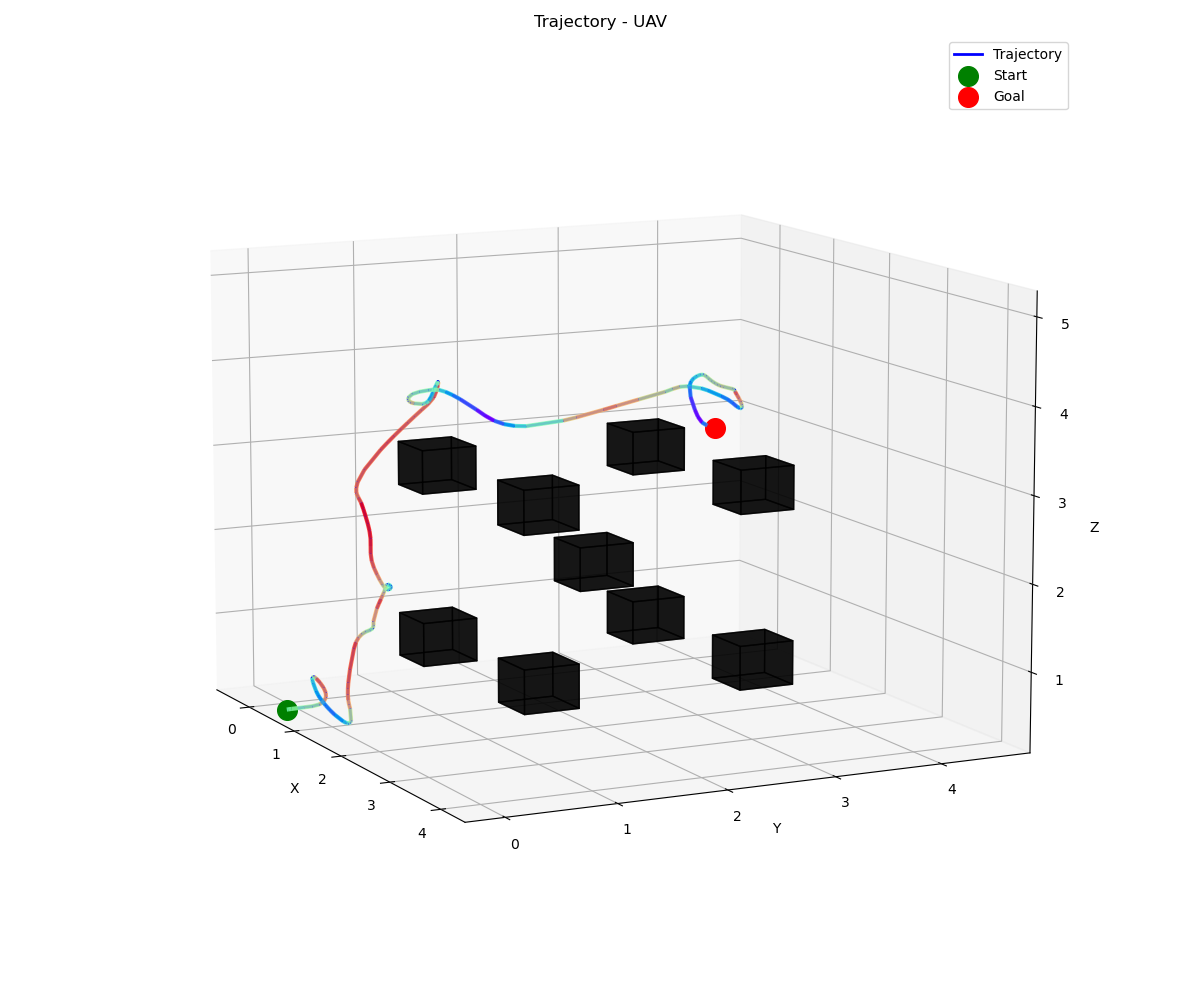}};
    \node[fill=white, opacity=0.85, text opacity=1, rounded corners, inner sep=2pt]
      at ([yshift=-2pt]img.south) {\textbf{Quad-1}};
  \end{tikzpicture}
}
& RRT\cite{lavalle2001randomized}        & 7.815 $\pm$ 5.281 & 7.558 $\pm$ 1.099 & \textbf{100.0} & 0.443 $\pm$ 0.088 & 0.0364 $\pm$ 0.0041 \\
\cline{2-7}
& SST\cite{li2016asymptotically}        & 30.197 $\pm$ 0.009 & 7.822 $\pm$ 1.126 & \textbf{100.0} & 0.479 $\pm$ 0.159 & 0.0374 $\pm$ 0.0063 \\
\cline{2-7}
& EST\cite{hsu2002randomized}        & 42.196 $\pm$ 1.998 & 8.027 $\pm$ 3.189 & \textbf{100.0} & 0.578 $\pm$ 0.040 & 0.0293 $\pm$ 0.0047 \\
\cline{2-7}
& KPIECE\cite{csucan2009kinodynamic}    & 29.366 $\pm$ 2.941 & 25.404 $\pm$ 27.838 & \textbf{100.0} & \textbf{1.769 $\pm$ 1.879} & 0.0460 $\pm$ 0.0074 \\
\cline{2-7}
& BOW\cite{raxit2025bow}        & \textbf{0.094 $\pm$ 0.027} & 6.673 $\pm$ 0.845 & \textbf{100.0} & 0.575 $\pm$ 0.132 & 0.0317 $\pm$ 0.0038 \\
\cline{2-7}
& BOWConnect & 0.117 $\pm$ 0.092 & \textbf{6.562 $\pm$ 0.790} & \textbf{100.0} & 0.568 $\pm$ 0.068 & \textbf{0.0279 $\pm$ 0.0039} \\

\midrule

\multirow{6}{*}{%
  \begin{tikzpicture}
    \node[inner sep=0] (img) {\includegraphics[width=0.145\textwidth, trim=120 100 85 140, clip]{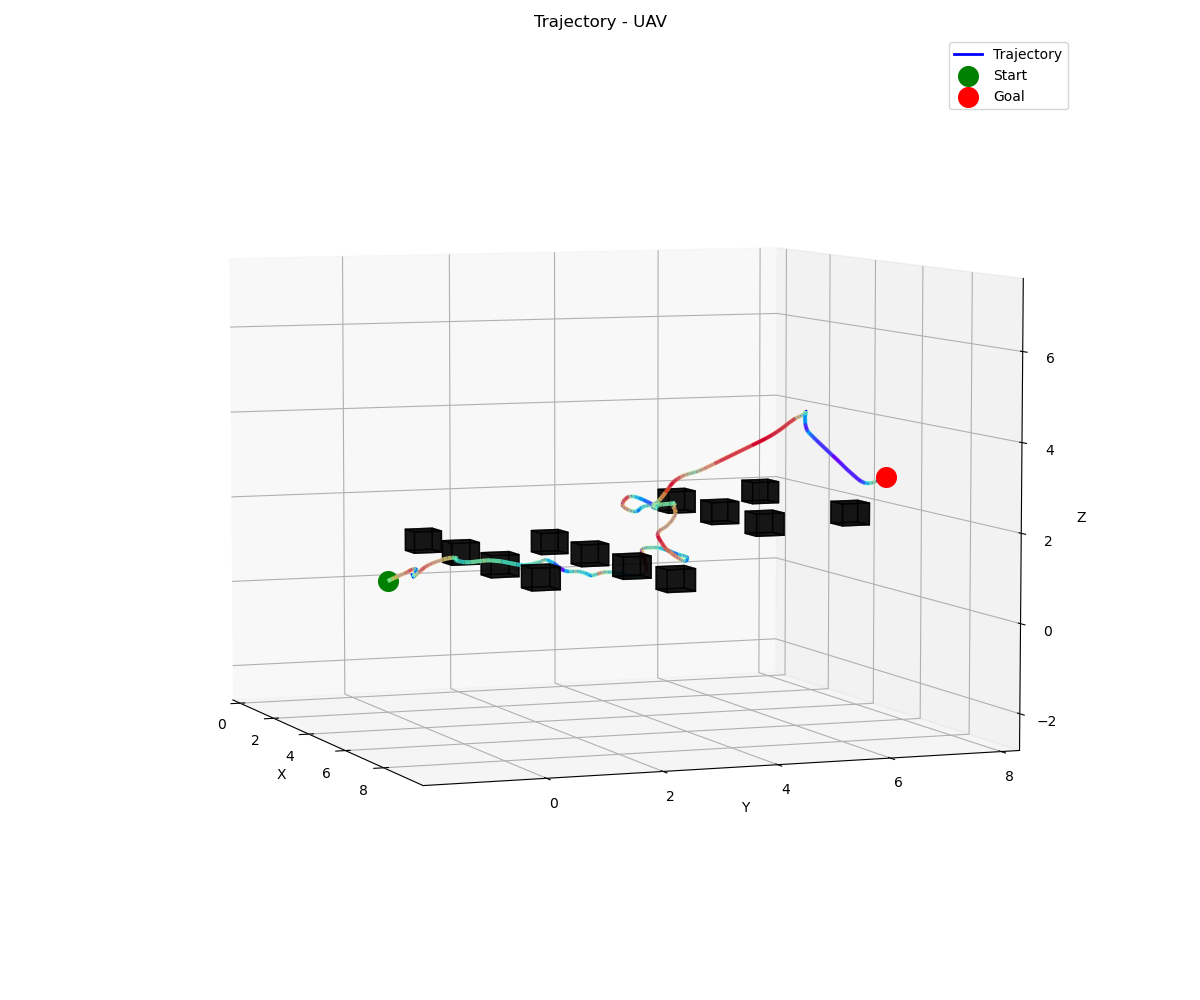}};
    \node[fill=white, opacity=0.85, text opacity=1, rounded corners, inner sep=2pt]
      at ([yshift=-2pt]img.south) {\textbf{Quad-2}};
  \end{tikzpicture}
}
& RRT\cite{lavalle2001randomized}        & 4.558 $\pm$ 3.380 & 16.835 $\pm$ 2.252 & \textbf{100.0} & 0.792 $\pm$ 0.170 & \textbf{0.0341 $\pm$ 0.0030} \\
\cline{2-7}
& SST\cite{li2016asymptotically}        & 30.200 $\pm$ 0.006 & \textbf{16.715 $\pm$ 2.075} & \textbf{100.0} & 0.886 $\pm$ 0.451 & 0.0353 $\pm$ 0.0028 \\
\cline{2-7}
& EST\cite{hsu2002randomized}        & 41.701 $\pm$ 0.989 & 25.492 $\pm$ 7.931 & \textbf{100.0} & \textbf{2.686 $\pm$ 2.178} & 0.0533 $\pm$ 0.0221 \\
\cline{2-7}
& KPIECE\cite{csucan2009kinodynamic}    & 32.116 $\pm$ 0.093 & 30.546 $\pm$ 16.021 & \textbf{100.0} & 2.155 $\pm$ 1.659 & 0.0519 $\pm$ 0.0179 \\
\cline{2-7}
& BOW\cite{raxit2025bow}        & 0.103 $\pm$ 0.040 & 33.944 $\pm$ 34.635 & \textbf{100.0} & 0.684 $\pm$ 0.066 & 0.0470 $\pm$ 0.0041 \\
\cline{2-7}
& BOWConnect & \textbf{0.086 $\pm$ 0.069} & 27.145 $\pm$ 26.096 & \textbf{100.0} & 0.680 $\pm$ 0.072 & 0.0432 $\pm$ 0.0053 \\
\midrule

\multirow{6}{*}{%
  \begin{tikzpicture}
    \node[inner sep=0] (img) {\includegraphics[width=0.145\textwidth, trim=120 100 85 140, clip]{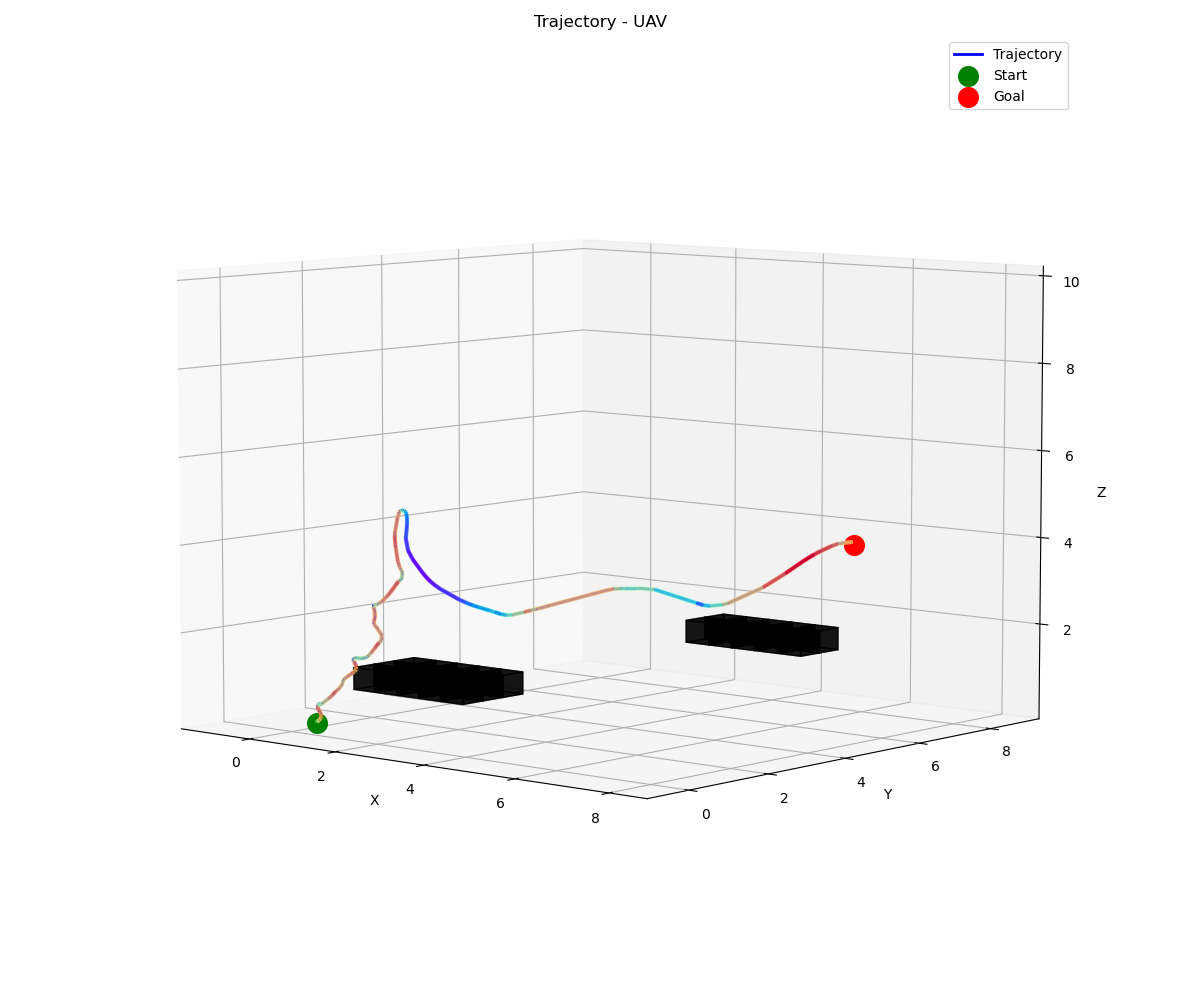}};
    \node[fill=white, opacity=0.85, text opacity=1, rounded corners, inner sep=2pt]
      at ([yshift=-2pt]img.south) {\textbf{Quad-3}};
  \end{tikzpicture}
}
& RRT\cite{lavalle2001randomized}        & 3.885 $\pm$ 2.856 & 18.224 $\pm$ 3.044 & \textbf{100.0} & 0.611 $\pm$ 0.230 & 0.0346 $\pm$ 0.0012 \\
\cline{2-7}
& SST\cite{li2016asymptotically}        & 30.208 $\pm$ 0.006 & \textbf{16.615 $\pm$ 0.492} & \textbf{100.0} & 0.775 $\pm$ 0.127 & \textbf{0.0334 $\pm$ 0.0031} \\
\cline{2-7}
& EST\cite{hsu2002randomized}        & 40.987 $\pm$ 0.695 & 16.943 $\pm$ 4.446 & \textbf{100.0} & 1.301 $\pm$ 0.662 & 0.0364 $\pm$ 0.0210 \\
\cline{2-7}
& KPIECE\cite{csucan2009kinodynamic}    & 32.116 $\pm$ 0.104 & 26.399 $\pm$ 16.319 & \textbf{100.0} & \textbf{2.269 $\pm$ 2.273} & 0.0445 $\pm$ 0.0110 \\
\cline{2-7}
& BOW\cite{raxit2025bow}        & \textbf{0.074 $\pm$ 0.009} & 22.674 $\pm$ 10.714 & \textbf{100.0} & 0.657 $\pm$ 0.039 & 0.0504 $\pm$ 0.0019 \\
\cline{2-7}
& BOWConnect & 0.081 $\pm$ 0.029 & 17.079 $\pm$ 2.566 & \textbf{100.0} & 0.705 $\pm$ 0.057 & 0.0467 $\pm$ 0.0062 \\
\midrule

\multirow{6}{*}{%
  \begin{tikzpicture}
    \node[inner sep=0] (img) {\includegraphics[width=0.145\textwidth, trim=120 100 85 140, clip]{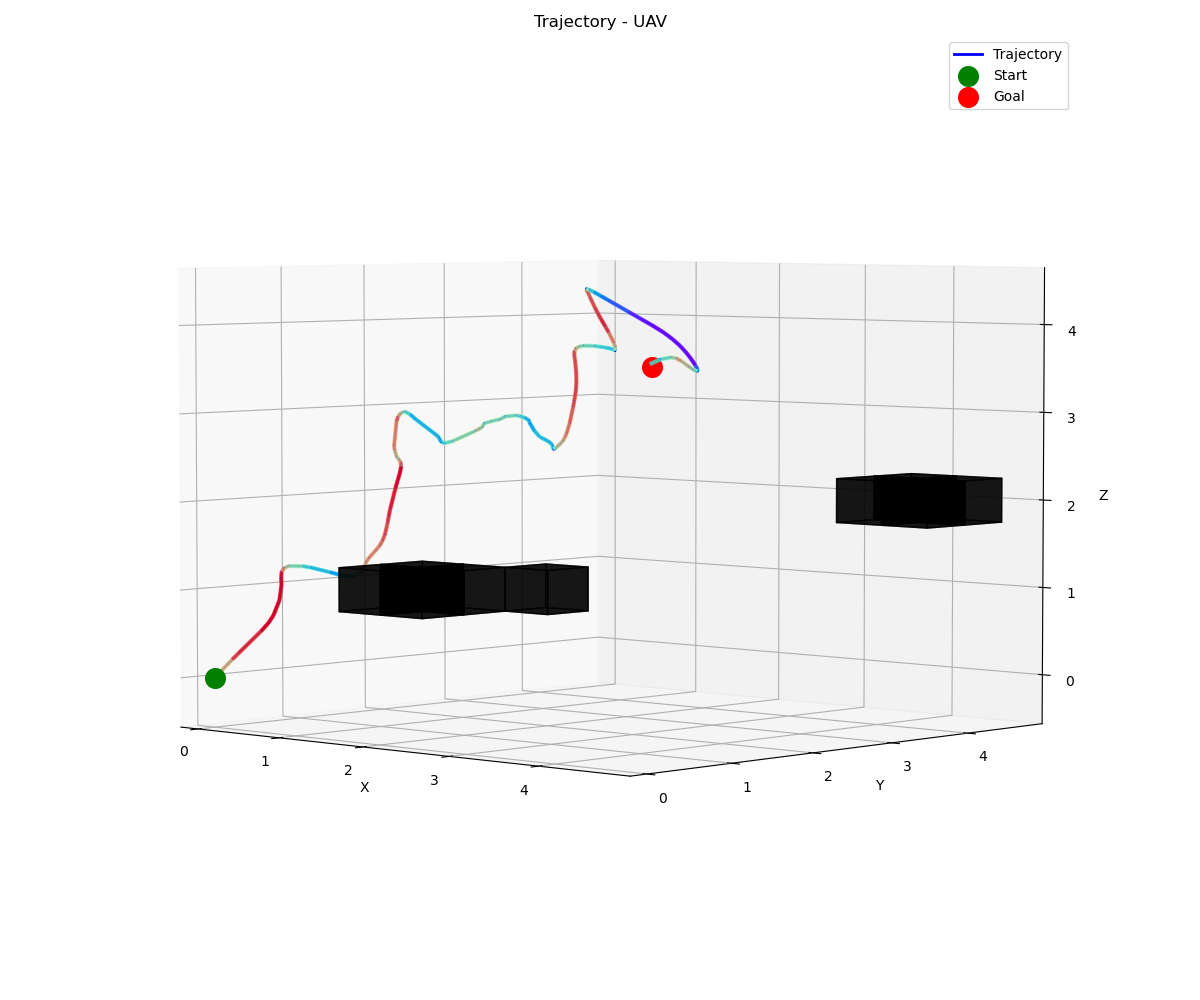}};
    \node[fill=white, opacity=0.85, text opacity=1, rounded corners, inner sep=2pt]
      at ([yshift=-2pt]img.south) {\textbf{Quad-4}};
  \end{tikzpicture}
}
& RRT\cite{lavalle2001randomized}        & 3.802 $\pm$ 2.566 & 9.866 $\pm$ 1.714 & \textbf{100.0} & 0.531 $\pm$ 0.129 & 0.0372 $\pm$ 0.0047 \\
\cline{2-7}
& SST\cite{li2016asymptotically}        & 30.198 $\pm$ 0.007 & 9.833 $\pm$ 3.309 & \textbf{100.0} & 0.644 $\pm$ 0.212 & 0.0372 $\pm$ 0.0056 \\
\cline{2-7}
& EST\cite{hsu2002randomized}        & 40.895 $\pm$ 1.017 & 11.502 $\pm$ 9.871 & \textbf{100.0} & \textbf{1.201 $\pm$ 0.668} & 0.0320 $\pm$ 0.0145 \\
\cline{2-7}
& KPIECE\cite{csucan2009kinodynamic}    & 31.475 $\pm$ 1.591 & 11.528 $\pm$ 3.423 & \textbf{100.0} & 0.819 $\pm$ 0.294 & 0.0426 $\pm$ 0.0032 \\
\cline{2-7}
& BOW\cite{raxit2025bow}        & \textbf{0.058 $\pm$ 0.014}& 6.463 $\pm$ 1.595 & \textbf{100.0} & 0.637 $\pm$ 0.107 & 0.0278 $\pm$ 0.0049 \\
\cline{2-7}
& BOWConnect & 0.091 $\pm$ 0.078 & \textbf{6.231 $\pm$ 0.572} & \textbf{100.0} & 0.597 $\pm$ 0.043 & \textbf{0.0246 $\pm$ 0.0048} \\

\bottomrule
\end{tabular}
\end{adjustbox}
\caption{Comparison of state-of-the-art kinodynamic motion planning methods across four quadrotor environments (Quad-1, Quad-2, Quad-3, and Quad-4). The best results in each environment are highlighted in \textbf{bold}. All planners were evaluated using a quadrotor motion model. Overlaid on each environment are qualitative BOWConnect trajectories, where color encodes the linear velocity profile.}
\label{tab:uav_results}
\vspace{-15pt}
\end{table*}

\subsection{UAV Benchmark Results}

We evaluate \textit{BOWConnect} across four quadrotor environments (\textit{Quad-1} through \textit{Quad-4}) using the same baselines and evaluation metrics (Table~\ref{tab:uav_results}). All planners achieve 100\% success in every environment. However, significant differences arise in computation time and trajectory quality.

BOWConnect and BOW both compute solutions in under $0.12$~s, while SST, EST, and KPIECE consistently require $30$--$42$~s. In \textit{Quad-1}, BOWConnect requires $0.117$~s compared to $7.815$~s for RRT and over $29$~s for SST and KPIECE. BOWConnect achieves the fastest time in \textit{Quad-2} ($0.086$~s), while BOW is slightly faster in \textit{Quad-1} ($0.094$~s), \textit{Quad-3} ($0.074$~s), and \textit{Quad-4} ($0.058$~s).

BOWConnect produces the shortest trajectories in \textit{Quad-1} ($6.562$~m) and \textit{Quad-4} ($6.231$~m), with consistently lower variance than BOW. In \textit{Quad-2} and \textit{Quad-3}, SST yields the shortest trajectories ($16.715$~m and $16.615$~m), but requires over $30$~s to compute them. BOWConnect also achieves the lowest jerk in \textit{Quad-1} ($0.0279$) and \textit{Quad-4} ($0.0246$), indicating smoother control profiles than classical planners while maintaining computation times two orders of magnitude lower.

\subsection{Unmanned Ground Vehicle Real-World Experiments}

\begin{figure}[htb]
    \centering
    \begin{subfigure}{0.88\columnwidth}
        \centering
        \includegraphics[width=\linewidth,trim = 0.0cm 3.75cm 0.0cm 3.75cm,
        clip]{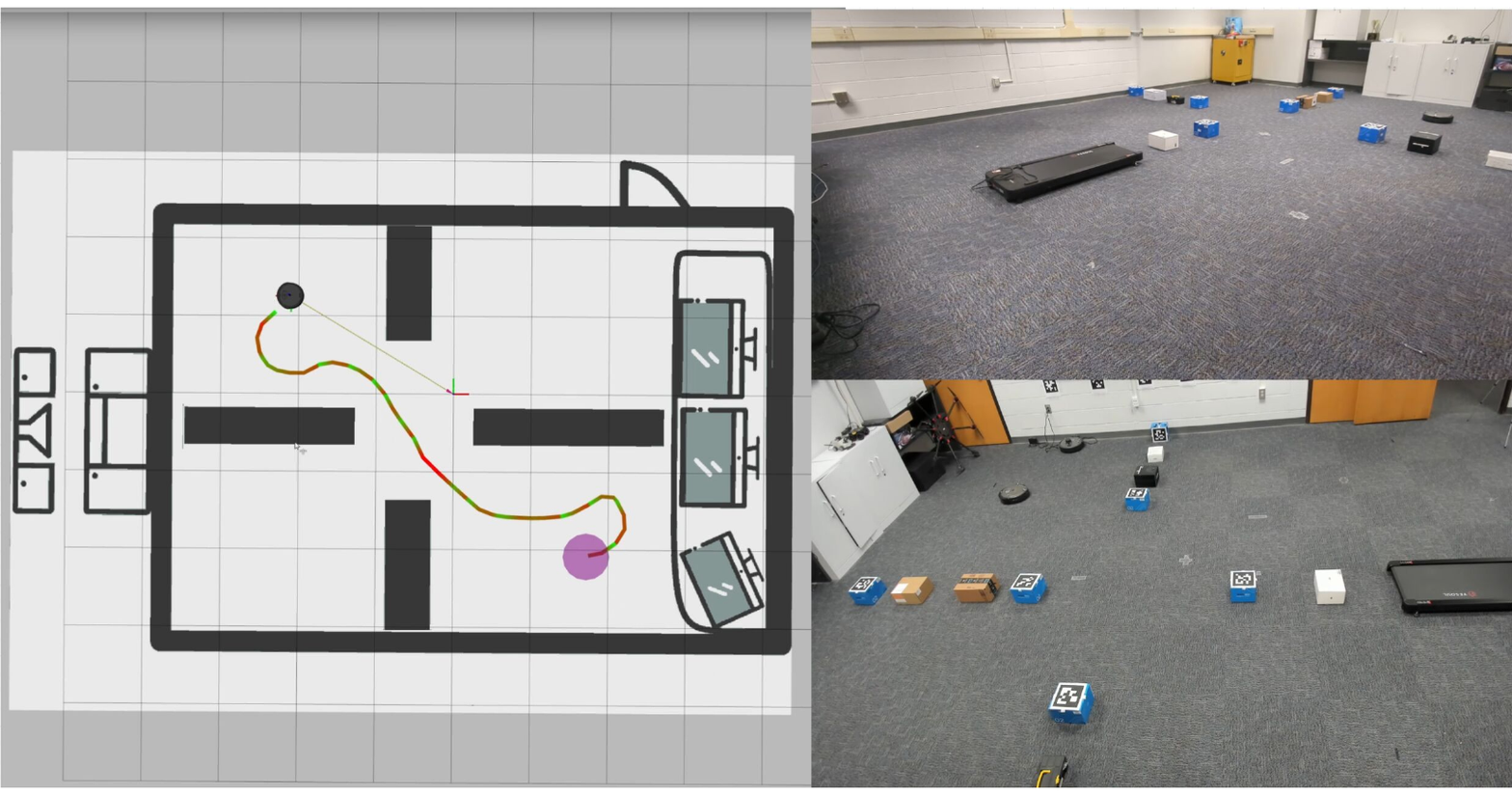}
        \caption{}
        \label{subfig:ugv_real_1}
    \end{subfigure}
    \hfill
    \begin{subfigure}{0.88\columnwidth}
        \centering
        \includegraphics[width=\linewidth,trim = 0.0cm 3.75cm 0.0cm 3.75cm,
        clip]{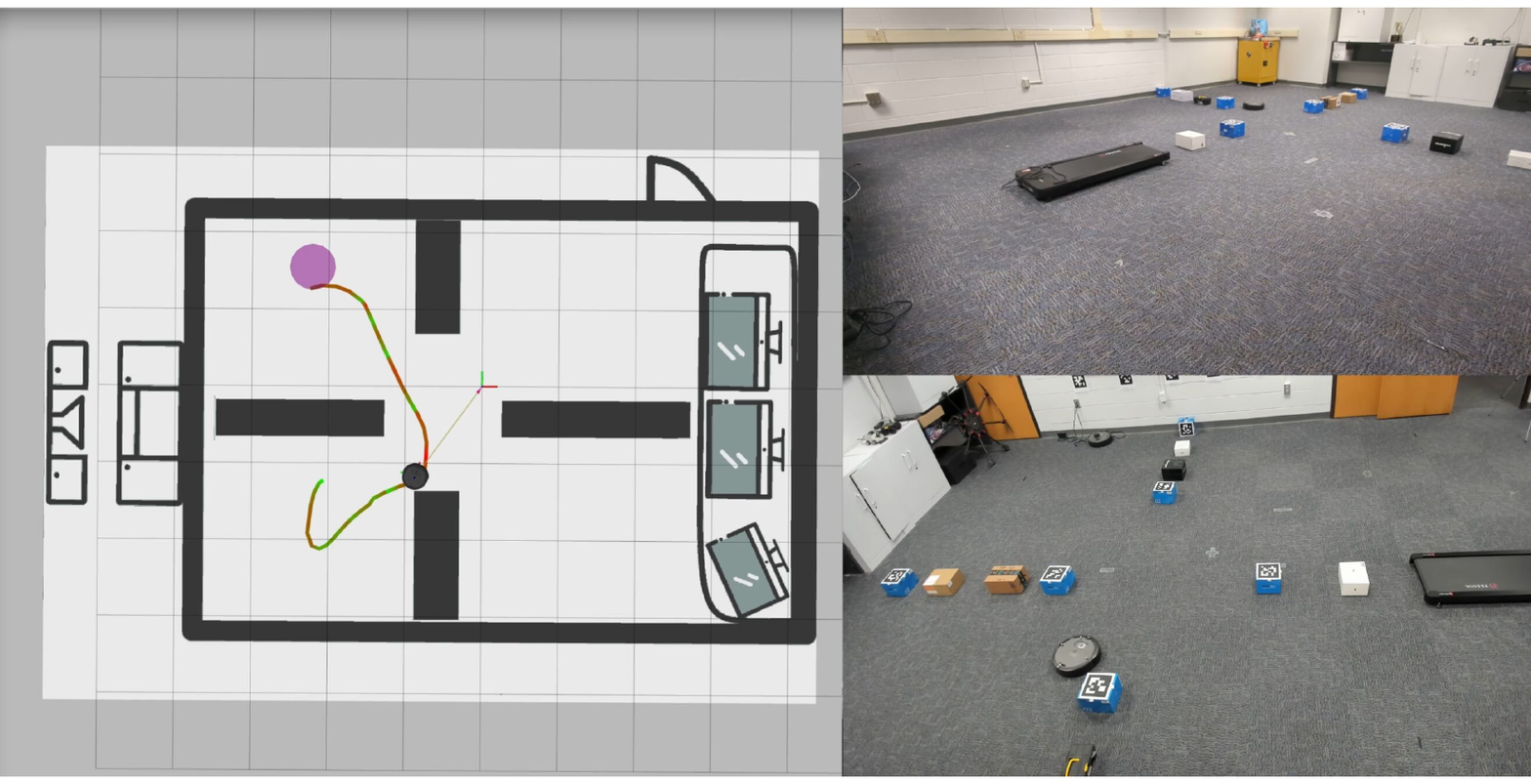}
        \caption{}
        \label{subfig:ugv_real_2}
    \end{subfigure}
    \caption{
    Experimental validation of the proposed kinodynamic motion planner in a cluttered indoor laboratory environment under two representative start–goal configurations. Fig.~\ref{subfig:ugv_real_1} and Fig.~\ref{subfig:ugv_real_2} illustrate the first and second configurations, respectively. In each case, the left panel shows the planned trajectory visualized in RViz, where the green-to-red color gradient indicates temporal progression, and the right panel shows the corresponding real-world UGV experiment conducted under similar obstacle layouts.
    }
    \label{fig:ugv_sim2real}
\end{figure}

We implemented BOWConnect on a non-holonomic differential-drive Create 3 educational robot with a radius of 0.17~m, operating within a bounded 6.5~m × 5.5~m workspace populated with obstacle configurations designed to emulate the constrained geometry of the benchmark scenario. Collision checking was performed using a 2D occupancy grid map constructed from the laboratory layout, while localization was provided by a Vicon motion capture system to ensure high-precision state estimation during execution. The UGV is modeled using a five-dimensional state vector $\mathbf{x} = (x, y, \theta, v, \omega)$, where $(x, y)$ denote planar position, $\theta$ is the heading angle, $v$ the linear velocity, and $\omega$ the angular velocity, with control input $\mathbf{u} = (v_c, \omega_c)$ representing commanded linear and angular velocities with respect to the body frame. The Unicycle motion model was used for the physical experiments. As illustrated in Fig.~\ref{fig:ugv_sim2real}, the goal pose (marked as a purple disk) was specified interactively in RViz using a 2D goal command, after which BOWConnect generated a dynamically feasible trajectory that was executed by the robot. The primary objective was to validate kinodynamic feasibility and real-time performance, and multiple start–goal configurations were tested across the workspace, with BOWConnect consistently generating feasible trajectories in under 0.15~s. During execution, the maximum linear velocity was limited to 1.0~m/s and the maximum angular velocity to 1.5~rad/s, while the low-level controller operated with a control time step of $\Delta t = 0.05$~s. Across five independent trials with varying start–goal configurations, no collisions or tracking instabilities were observed during execution. The average planning time was 0.12~s, with a maximum of 0.15~s across all runs, demonstrating real-time feasibility on the onboard compute platform.

\begin{figure}[t]
    \centering
    \begin{subfigure}{0.9\columnwidth}
        \centering
        \includegraphics[width=\linewidth,trim = 0.0cm 3.75cm 0.0cm 3.75cm,
        clip]{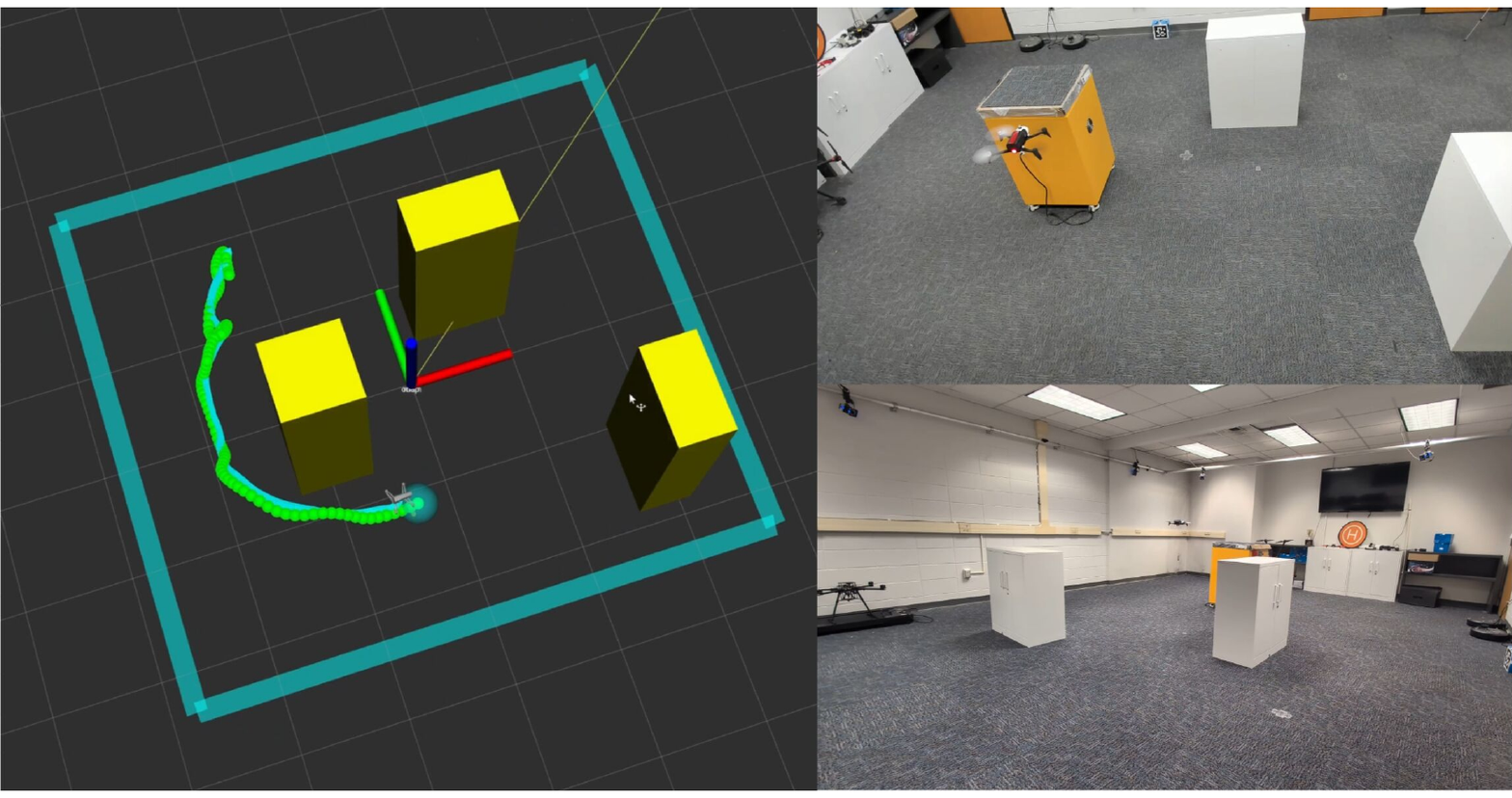}
        \caption{}
        \label{subfig:uav_real_1}
    \end{subfigure}
    \hfill
    \begin{subfigure}{0.9\columnwidth}
        \centering
        \includegraphics[width=\linewidth,trim = 0.0cm 3.75cm 0.0cm 3.75cm,
        clip]{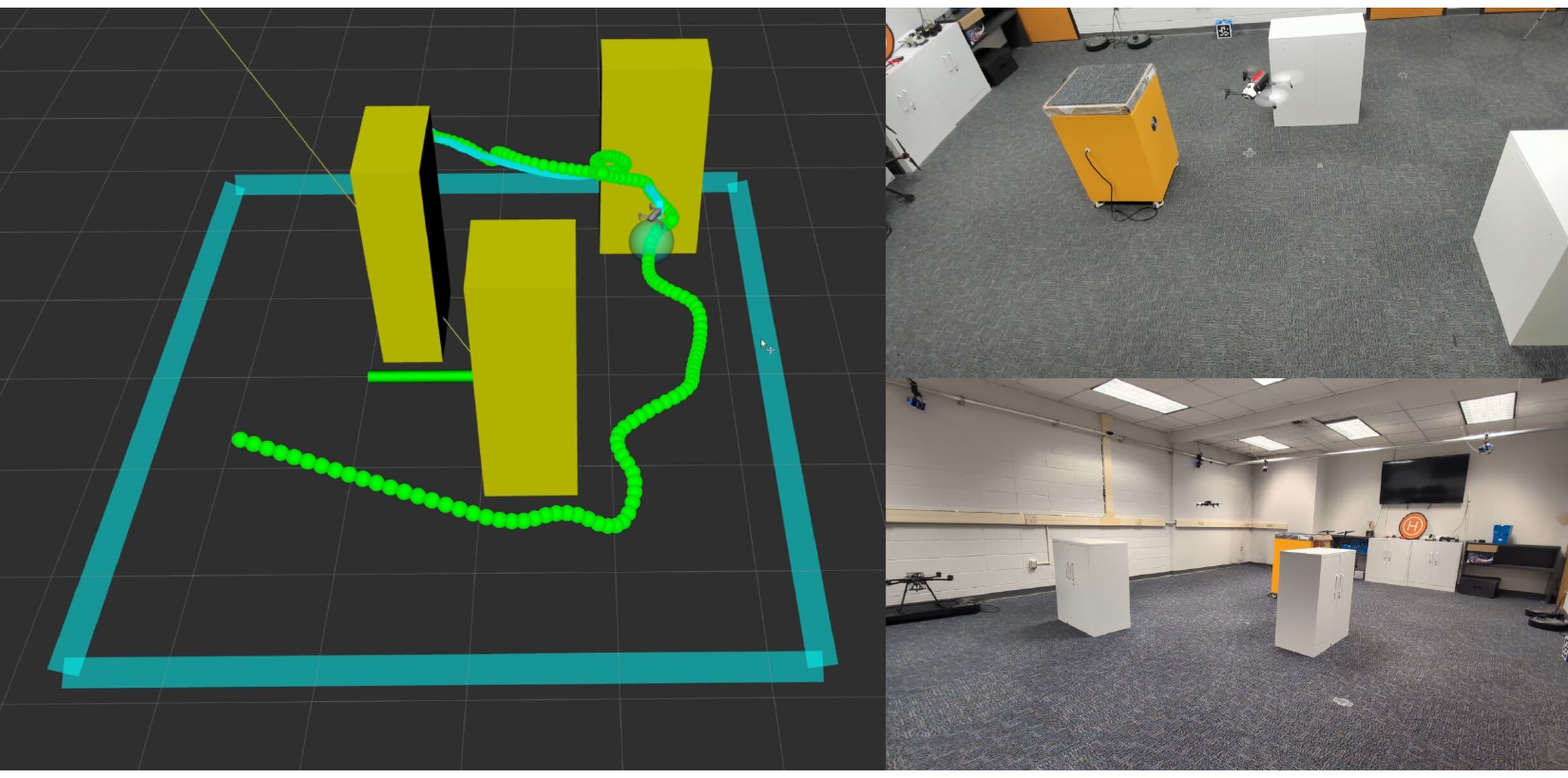}
        \caption{}
        \label{subfig:uav_real_2}
    \end{subfigure}
    \caption{
    BOWConnect quadrotor experiments in a cluttered indoor environment with a fixed obstacle layout and two distinct start–goal configurations. Fig.~\ref{subfig:uav_real_1} shows the first configuration, while Fig.~\ref{subfig:uav_real_2} shows the second configuration. In each case, the left panel presents the simulated planning setup with the generated kinodynamic trajectory (green) visualized in RViz, and the right panel shows the corresponding real-world quadrotor experiment conducted under the identical obstacle arrangement, demonstrating sim-to-real consistency in trajectory execution.
    }
    \label{fig:uav_sim2real}
\end{figure}

\subsection{Unmanned Aireal Vehicle Physical Experiments}

We implemented \textit{BOWConnect} on a 6-DOF quadrotor UAV in a three-dimensional environment with 3D obstacles to demonstrate the capability of the proposed planner in high-dimensional planning problems, as shown in Fig.~\ref{fig:uav_sim2real}. The experiments were conducted using a Parrot Bebop~2 UAV. The UAV state is modeled using an 8-dimensional vector $\mathbf{x} = (x, y, z, \theta, \dot{x}, \dot{y}, \dot{z}, \dot{\theta})$, which includes position, yaw orientation, and linear and angular velocities. The control input is defined as a 4-dimensional vector $\mathbf{u} = (\dot{x}_c, \dot{y}_c, \dot{z}_c, \dot{\theta}_c)$, representing commanded body-frame linear velocities and yaw rate. In Fig.~\ref{fig:uav_sim2real}, the left panels show the RViz visualization with the real UAV state and the planned trajectory generated by BOWConnect (green), while the cyan curve denotes the actual trajectory followed by the UAV during execution. The goal position is sent interactively from RViz using the \textit{Nav Goal} tool. The experiments were conducted in a bounded workspace of $6.5~\text{m} \times 5.5~\text{m} \times 2.5~\text{m}$ with three box-shaped obstacles placed to replicate the RViz configuration: two identical obstacles of dimensions $0.4~\text{m} \times 0.8~\text{m} \times 0.92~\text{m}$ and one obstacle of dimensions $0.64~\text{m} \times 0.64~\text{m} \times 1.08~\text{m}$. The close alignment between the planned trajectory (green) and the executed trajectory (cyan) demonstrates accurate tracking and confirms the dynamic feasibility of the generated paths. Multiple trials were conducted from different start and goal configurations, and in all cases BOWConnect consistently computed collision-free trajectories in under $0.1~\text{s}$, demonstrating computational efficiency and robustness in high-dimensional kinodynamic planning.

    \section{Conclusion}
This paper presented BOWConnect, a bidirectional parallel kinodynamic motion planner that combines parallel Bayesian Optimization over Windows local planners with spatial hashing and boundary value solver to overcome the sample inefficiency, heuristic limitations, and narrow passage failures of existing kinodynamic planners. BOWConnect replaces random control sampling with GP-guided acquisition and grows trees from both start and goal regions in parallel. Across ten benchmark environments evaluated under both kinematic and dynamic motion models, BOWConnect consistently achieves 100\% success and the fastest or near-fastest computation time, typically solving problems in under 0.019s to 2.133s while classical planners such as SST, EST, and KPIECE frequently require 7 to 42~s or fail entirely. Real-world experiments on a ground vehicle and a quadrotor confirm real-time performance under 0.15~s with no collisions, validating the practical applicability of the approach. Future work will explore adaptive worker allocation, planning under uncertainty, and extension to dynamic environments with moving obstacles.

    \begingroup
    \footnotesize
    \bibliography{main,bowconnect}
    \bibliographystyle{IEEEtran}
    \endgroup

\end{document}